\documentclass[a4paper,fleqn]{cas-dc}

\usepackage{amssymb}
\usepackage{balance}
\usepackage{amsmath}
\usepackage{graphicx}
\usepackage{hyperref}
\usepackage[linesnumbered,ruled]{algorithm2e}
\usepackage{xcolor}
\usepackage{soul}
\usepackage{hhline}
\usepackage[normalem]{ulem}
\usepackage{stmaryrd}

 \usepackage{multirow}
 \usepackage{multicol}
 \usepackage{booktabs}
 \usepackage{fancyhdr}
 \usepackage{tabularx}
 \usepackage{float}
\usepackage{comment}
\usepackage{arydshln}
\usepackage{color}
\usepackage{geometry}
\definecolor{columbiablue}{rgb}{0.61, 0.87, 1.0}
\definecolor{battleshipgrey}{rgb}{0.52, 0.52, 0.51}
\definecolor{davygrey}{rgb}{0.33, 0.33, 0.33}
\definecolor{denim}{rgb}{0.08, 0.38, 0.74}

\usepackage{lineno}
\usepackage{appendix}

\usepackage[numbers]{natbib}
\def\tsc#1{\csdef{#1}{\textsc{\lowercase{#1}}\xspace}}
\tsc{WGM}
\tsc{QE}
\tsc{EP}
\tsc{PMS}
\tsc{BEC}
\tsc{DE}
\begin{document}
\let\WriteBookmarks\relax
\def\floatpagepagefraction{1}
\def\textpagefraction{.001}

\shorttitle{Simplifying approach to Node Classification in Graph Neural Networks}

\shortauthors{Maurya et~al.}

%To remove Preprint submitted to Elsevier
\ExplSyntaxOn
 \cs_gset:Npn \__first_footerline:
   { \group_begin: \small \sffamily \__short_authors: \group_end: }
 \ExplSyntaxOff

\title [mode = title]{Simplifying approach to Node Classification in Graph Neural Networks}                      

% \tnotetext[1]{This document is the results of the research
%   project funded by the National Science Foundation.}

% \tnotetext[2]{The second title footnote which is a longer text matter
%   to fill through the whole text width and overflow into
%   another line in the footnotes area of the first page.}

\let\printorcid\relax

\author[tokyotech,oil]{Sunil Kumar Maurya\corref{cor}}
\ead{skmaurya@net.c.titech.ac.jp}

\cormark[1]

\author[airc,oil]{Xin Liu}
\ead{xin.liu@aist.go.jp}

\author[tokyotech,oil]{Tsuyoshi Murata}
\ead{murata@c.titech.ac.jp}

\affiliation[tokyotech]{organization={Department of Computer Science, Tokyo Institute of Technology},%Department and Organization
            %addressline={}, 
            city={Tokyo},
            %postcode={}, 
            %state={},
            country={Japan}}

\affiliation[airc]{organization={Artificial Intelligence Research Center, AIST},%Department and Organization
            %addressline={}, 
            city={Tokyo},
            %postcode={}, 
            %state={},
            country={Japan}}

\affiliation[oil]{organization={AIST-Tokyo Tech Real World Big-Data Computation Open Innovation Laboratory},%Department and Organization
            %addressline={}, 
            city={Tokyo},
            %postcode={}, 
            %state={},
            country={Japan}}

\cortext[cor1]{Corresponding author}
% \cortext[cor2]{Principal corresponding author}

% \fntext[fn1]{This is the first author footnote. but is common to third
% \fntext[fn2]{Another author footnote, this is a very long footnote and
%   it should be a really long footnote. But this footnote is not yet
%   sufficiently long enough to make two lines of footnote text.}

% \nonumnote{This note has no numbers. In this work we demonstrate $a_b$
%   the formation Y\_1 of a new type of polariton on the interface
%   between a cuprous oxide slab and a polystyrene micro-sphere placed

\begin{abstract}
Graph Neural Networks (GNNs) have become one of the indispensable tools to learn from graph-structured data, and their usefulness has been shown in wide variety of tasks. In recent years, there have been tremendous improvements in architecture design, resulting in better performance on various prediction tasks. In general, these neural architectures combine node feature aggregation and feature transformation using learnable weight matrix in the same layer. This makes it challenging to analyze the importance of node features aggregated from various hops and the expressiveness of the neural network layers. As different graph datasets show varying levels of homophily and heterophily in features and class label distribution, it becomes essential to understand which features are important for the prediction tasks without any prior information. In this work, we decouple the node feature aggregation step and depth of graph neural network, and empirically analyze how different aggregated features play a role in prediction performance. We show that not all features generated via aggregation steps are useful, and often using these less informative features can be detrimental to the performance of the GNN model. Through our experiments, we show that learning certain subsets of these features can lead to better performance on wide variety of datasets. Based on our observations, we introduce several key design strategies for graph neural networks. More specifically, we propose to use softmax as a regularizer and "soft-selector" of features aggregated from neighbors at different hop distances; and L2-Normalization over GNN layers. Combining these techniques, we present a simple and shallow model, Feature Selection Graph Neural Network (FSGNN), and show empirically that the proposed model achieves comparable or even higher accuracy than state-of-the-art GNN models in nine benchmark datasets for the node classification task, with remarkable improvements up to 51.1\%. Source code available at \url{https://github.com/sunilkmaurya/FSGNN/}

\end{abstract}

% \includegraphics{figs/grabs.pdf}

% \begin{highlights}
% \item Current Graph Neural Networks (GNNs) have inconsistent performance in homophily and heterophily graphs.
% \item We analyze importance of feature selection over hops with extensive experiments.
% \item With good feature selection strategy, simple NN model is sufficient for high accuracy.
% \item We propose a model FSGNN for node classification task.
% \item FSGNN outperforms SOTA on heterophily datasets.
% \item FSGNN provides comparable performance with SOTA GNN models in homophily graphs.
% \end{highlights}

% Keywords

\maketitle

\section{Introduction}
\label{introduction}

Graph Neural Networks (GNNs) have opened a unique path to learning on data by leveraging the intrinsic relations between entities that can be structured as a graph. By imposing these structural constraints, additional information can be learned and used for many types of prediction tasks. With rapid development of the field and easy accessibility of computation and data, GNNs have been used to solve a variety of problems like node classification \cite{kipf_semi-supervised_2017,velickovic_graph_2018,abu-el-haija_mixhop_2019,chen_simple_2020,wang_tree_2021}, link prediction \cite{ying_graph_2018,van_den_berg_graph_2017,chami_hyperbolic_2019}, graph classification \cite{ying_hierarchical_2018,zhang_end--end_2018}, prediction of molecular properties \cite{gilmer_neural_2017,madhawa_graphnvp_2019}, node ranking \cite{maurya_graph_2021,fan_learning_2019} and natural language processing \cite{marcheggiani_encoding_2017}.

In this work, we focus on the node classification task using graph neural networks. Since the success of early GNN models such as GCN \cite{kipf_semi-supervised_2017}, researchers have successively proposed numerous variants \cite{wu_comprehensive_2019} to address its various shortcomings in model training and to improve the prediction capabilities. Some of the techniques used in these variants include  neighbor sampling \cite{hamilton_inductive_2017,chen_fastgcn_2018}, attention mechanism \cite{velickovic_graph_2018}, use of Personalized PageRank matrix instead of adjacency matrix \cite{klicpera_predict_2018}, leveraging proximity in feature space \cite{jin_node_2021} and simplified model design \cite{wu_simplifying_2019}. Also, there has been an increasing trend in making the models deeper by stacking more layers and using the residual connections to improve the expressiveness of the model \cite{rong_dropedge_2020,chen_simple_2020}. For example, GCNII \cite{chen_simple_2020} incorporates up to 64 layers with scaled residual weights.
However, most of these models by design are more suitable for homophily datasets, where nodes that  are linked to each other are more likely to belong to the same class. As a result, these GNNs may not perform well with heterophily datasets, which are more likely to have nodes with different labels connected together. This problem was highlighted by Zhu et al. \cite{zhu_beyond_2020} and the authors proposed node's ego-embedding and neighbor-embedding separation to improve performance on heterophily datasets. Other recent works approach this problem in different manner e.g, \cite{zhu_graph_2021} utilizes belief propagation, \cite{bo_beyond_2021} uses adaptive gating on edges etc.

In general, GNN models combine feature aggregation and transformation using a learnable weight matrix in the same layer, often referred to as graph convolutional layer.
These layers are stacked together with the non-linear transformation (e.g., ReLU) and regularization (e.g., Dropout) as a learning framework on the graph data. Stacking the layers also has the effect of introducing powers of adjacency matrix (or laplacian matrix), which helps to generate a new set of features for a node by aggregating neighbor's features at multiple hops, thus encoding the neighborhood information. The number of these unique features depends on the propagation steps or the depth of the model. The final node embeddings are the output of just stacked layers or, for some models, also have skip connection or residual connection combined at the final layer.

However, such a combination muddles the distinction between the importance of features and the expressiveness of Multi-layer Perceptron (MLP). It becomes challenging to analyze which component contributes more over a specific prediction task. In this paper, we treat feature propagation and learning on neural network separately and run extensive experiments to study the importance of features in improving the prediction capabilities of the model. Based on our analysis, we propose a simple GNN model to improve prediction accuracy in the node classification task.

\noindent
\textbf{Our Contributions}\hfill

\begin{itemize}
    \item We run extensive experiments on multiple node classification benchmark datasets and show that hop feature
selection is an essential requirement for higher prediction accuracy.
    \item We experimentally confirm the feature generation requirement for homophily and heterophily graphs.
    \item We propose a simple 2-layered GNN model  \textbf{FSGNN}\footnote{This work is an extension of our previous work \cite{maurya_improving_2021}.}, which incorporates a "soft-selection" mechanism
to learn the importance of features during training of the model.
    \item Our proposed model empirically outperforms other state-of-art (SOTA) GNN models (both shallow and deep) and achieves up to 51.1\% higher node classification accuracy.

\end{itemize}

% In addition, analyzing the model parameters gives us an insight into identifying which features are most responsible
% for classification accuracy. One interesting observation we find is regarding Chameleon and Squirrel datasets. These are dense graph datasets and are generally regarded as being low-quality heterophily datasets. However, in our experiments with our proposed model, we find them to be showing strong heterophily properties with improved classification results. 

The rest of the paper is organized as follows: Section \ref{preliminaries} outlines formulation of graph neural networks and details node classification task. In Section \ref{sec:feat_imp}, we explore the importance of features by running extensive experiments and present our observations. Based on our experimental observations, in Section \ref{propose_arch}, we propose our GNN model FSGNN. In Section \ref{related_work}, we briefly introduce relevant GNN literature. Section \ref{experiments} contains the experimental details and comparison with other GNN models. In Section \ref{sec:results}, we present our results and empirically analyze our proposed design strategies and their effect on the model's performance. Section \ref{conclusion} summarizes the paper.

%\vspace{-1mm}
\section{Preliminaries}
\label{preliminaries}

Let $G = (V,E)$ be an undirected graph with $n$ nodes and $m$ edges. For numerical calculations, graph is represented as adjacency matrix denoted by $A\in \{0,1\}^{n\times n}$ with each element $A_{ij}=1$ if there exists an edge between node $v_i$ and $v_j$, otherwise $A_{ij}=0$. When self-loops are added to the graph then, resultant adajcency matrix is denoted as $\Tilde{A} = A+I$. Diagonal degree matrix of $A$ and $\Tilde{A}$ are denoted as $D$ and $\Tilde{D}$ respectively. Each node is associated with a d-dimensional feature vector and the feature matrix for all nodes is represented as $X \in \mathbb{R} ^{n \times d}$. In our discussion, we will interchangeably refer to feature matrix as features of the nodes.

\subsection{Graph Neural Networks}
\label{sec:gnn}

GNNs leverage feature propagation mechanism \cite{gilmer_neural_2017} to aggregate neighborhood information of a node and use non-linear transformation with trainable weight matrix to get the final embeddings for the nodes. Conventionally, a simple GNN layer is defined as 
\begin{equation}
  \label{eq:homophily_gnn}
  H^{(i+1)} = \sigma (\Tilde{A}_{sym}H^{(i)}W^{(i)})
\end{equation}

\noindent
where $\Tilde{A}_{sym} = \Tilde{D}^{-\frac{1}{2}} \Tilde{A}\Tilde{D}^{-\frac{1}{2}}$ is a symmetric normalized adjacency matrix with added self-loops. $H^{(i)}$ represents features from the previous layer, $W^{(i)}$ denotes the learnable weight matrix, and $\sigma$ is a non-linear activation function, which is usually ReLU in most implementations of GNNs.  However, this formulation is suitable for homophily datasets as features are cumulatively aggregated, i.e. node's own features are added together with neighbor's features. The cumulative aggregation of node's self-features with that of neighbors reinforces the signal corresponding to the label and helps to improve accuracy of the predictions. On the other hand, in the case of heterophily, nodes are assumed to have dissimilar features and labels to their neighbors. For heterophily datasets, H2GCN \cite{zhu_beyond_2020} proposes to separate features of neighbors from node's own features, thus avoiding aggregation of dissimilar features. So we use the following formulation for the GNN layer,

\begin{equation}
      \label{eq:heterophily_gnn}
      H^{(i+1)} = \sigma (A_{sym}H^{(i)}W^{(i)})
\end{equation}
where $A_{sym} = D^{-\frac{1}{2}} AD^{-\frac{1}{2}}$ is symmetric normalized adjacency matrix without added self-loops. To combine features from multiple hops, concatenation operator can be used before the final layer.

Following the conventional GNN formulation \cite{kipf_semi-supervised_2017} using $\Tilde{A}$, a simple 2-layered GNN can be represented as,

\begin{equation}
    \label{eq:gnn_2layer}
    Z = \Tilde{A}_{sym}\sigma(\Tilde{A}_{sym}XW^{(0)})W^{(1)}
\end{equation}

\begin{figure}
    \centering
    \includegraphics[width=\linewidth]{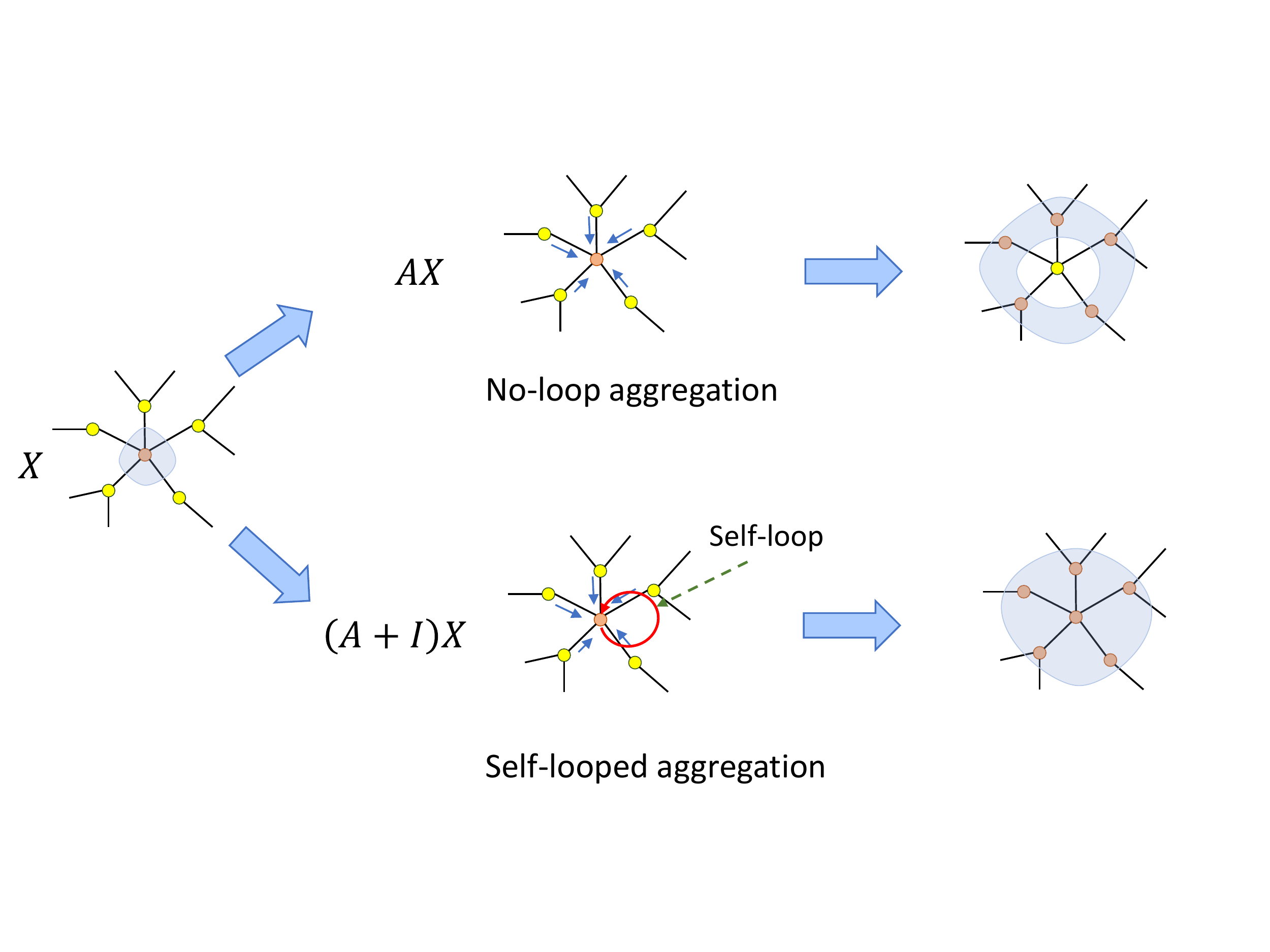}
    \vspace{-1cm}
    \caption{No-loop and Self-loop aggregation of node features}
    \label{fig:scalar_val_heat}
\end{figure}

\subsection{Node Classification}

Node classification is an extensively studied graph based semi-supervised learning problem. It encompasses training the GNN to predict labels of nodes based on the features and neighborhood structure of the nodes. GNN model is considered as a function $f(X,A)$ conditioned on node features $X$ and adjacency matrix $A$. Using Eq. (\ref{eq:gnn_2layer}), GNN aggregates the features of two hops of neighbors and outputs $Z$. Softmax function is applied row-wise, and cross-entropy error is calculated over all labeled training examples. The gradients of loss are back-propagated through the GNN layers. Once trained, the model can be used to predict labels of nodes in the test set.

\subsection{Feature Generation in GNNs}
As discussed in section \ref{sec:gnn}, we can generate different features for the nodes capturing homophilic and heterophilic properties in the graph by modifying the feature aggregation step. The number of different features generated is dependent on the number of hops of aggregation over neighbors. In addition, node features can also be generated based on proximity in feature space or based on some other arbitrary criterion. Feature generation steps can differ among GNNs based on their architecture. Many GNN models have feature aggregation and representation learning combined in single layer \cite{kipf_semi-supervised_2017,klicpera_predict_2018,chen_simple_2020}, while in other models features can be precomputed beforehand \cite{wu_simplifying_2019,frasca_sign_2020}.

\section{Feature Selection in Graph Neural Networks}
\label{sec:feat_imp}

% For the design of a GNN with good generalization capability and performance, there are many aspects of the data that need to be considered. The feature propagation and aggregation scheme is governed by if the class label distribution has strong homophily or heterophily or some combination of both. The number of hops (and depth of the model for many GNN models) for feature aggregation are dependent on graph structure and size as well as label distribution among neighbors of the nodes. Also, the type and amount of regularization during training needs to be decided, for example, using dropout on input features or on graph edges.

On any given graph-structured data, a set of features can be generated for the nodes (e.g. using Eq. (1) \& (2)). The number of features depends on the problem in hand, properties of the dataset, design choice of practitioner etc. 

We assume a function, 
$$g(X,A,K) \mapsto \{X_1,X_2, \dotsc ,X_l \}$$
The function takes $X \in \mathbb{R} ^{n \times d}$ as node features matrix, $A$ as an adjacency matrix, $K$ as the power of the adjacency matrix or number of hops to propagate features and outputs a set of $l$ node features.

However, in the node classification task, for given label distribution, only a subset of these features are relevant to predict the label of the node. For example, a feature $X_i$ is relevant to class $C_i$, if $X_i$ and $C_i$ are highly correlated\cite{tang_feature_2014,li_feature_2017,chandrashekar_survey_2014}. Irrelevant or noisy features may not correlate with target labels but can still affect the learning process.

In this section, we explore the importance of features generated by the aggregation step at different hops. We run a series of extensive experiments to study how different features affect predictions for graph neural networks in the node classification task. Using these experiments, we aim to answer the following three questions:

\textbf{Q.1} \textit{How useful are individual features generated from multi-step aggregation in graph neural networks?}

\textbf{Q.2} \textit{What is the effect of training the model over all the features, and what are the effects of different aggregator schemes?}

\textbf{Q.3} \textit{ What is the impact of adding or removing features on the model's performance?}

Exploring these questions provides a deeper understanding of how GNN models can be designed to have better prediction capabilities.

\subsection{Experiment Setting}

\subsubsection{Model Design}
\label{sec:design_strategy}
Conventionally, GNN models have feature propagation and transformation combined into a single layer, and the layers are stacked together. This step makes it difficult to distinguish the importance of the features and the role of MLP, and it becomes harder to analyze their impact on the prediction results of the model. For our experiments, we decouple the feature generation step and representation learning over features separately.

%This provides us with three main benefits.
% \renewcommand{\labelenumi}{(\roman{enumi})}
% \begin{enumerate}
%     \item Features generated for nodes are not constrained by the design of the GNN model. We get the freedom to choose the feature set as required by the problem and the neural network design, which is sufficiently expressive.
%     \item We can precompute and fix the node features set and experiment with the neural network architectures for the best performance. Precomputing features also helps to scale the training of the model for large graphs with batchwise training.
%     \item In conventional GNN setting, stacking many layers also causes oversmoothing of node features \cite{chen_measuring_2019} and adversely affects the performance of the model. Recently proposed models use skip connection or residual connection to overcome this issue. However, they fail to demonstrate which features are useful. We provide an alternative scheme where the model can learn weights that identify which features are useful for the prediction task.
% \end{enumerate}

We use 2-layer neural network for our experiments.
For the model design, instead of a single input channel, we propose to have all these features as input in parallel. Each feature is mapped to a separate linear layer. Hence the linear transformations are uniquely learned for all input features. In addition, we use L2-normalization to row-wise normalize the output of the linear layer. L2-normalization scales the node embedding vectors to lie on the "unit sphere". \texttt{ReLU} and \texttt{Dropout} are used for non-linear transformation of hidden features and regularization respectively. In the case of a single input feature matrix, hidden features are mapped to the final layer, and in the case of multiple input features, all hidden features of the node are aggregated and mapped to the final layer. We use two different aggregator schemes: \texttt{sum} and \texttt{concatenation}, and compare the results.

\subsubsection{Input Features}

In our experiments, we consider node features of up to 3-hop neighbors (commonly used setting) in the graph. As we analyze both heterophily and homophily properties in a graph, we calculate both self-looped and no-loop features of the nodes. Hence, including node's own features, our feature set has total of 7 different features for the node $X_{feat} = \{ X,AX, (A+I)X, A^2X, (A+I)^2X, A^3X, (A+I)^3X\}$. To explore the answers to the three questions defined earlier, we design three experiment settings: \texttt{Single\_feature},  \texttt{All\_feature} and \texttt{Sub\_feature}. In \texttt{Single\_feature} setting, we use only one out of seven features to train the model, and results are compared among all features. In this way, we analyze how informative each feature is for the label prediction of the nodes. In \texttt{All\_feature} setting, we train the model on all features together and ascertain the model's performance.  In addition, we use two aggregation schemes, i.e. sum and concatenation of hidden node features as they are commonly used in GNN models. Please note that \texttt{All\_feature} with concatenate setting is similar to the SIGN model\cite{frasca_sign_2020} as the model uses simple concatenation of features. In \texttt{Sub\_feature} setting, we train the model on all possible combinations of the features in $X_{feat}$ excluding the subsets already used in \texttt{Single\_feature} and \texttt{All\_feature}. Hence in this setting, we train the model on 119 different subsets of input features and report the result for the best one. We use both aggregator schemes in this case too.

\subsubsection{Datasets and Hyperparameters}

We run experiments on nine different datasets with varying homophily and heterophily properties. More details on datasets and preprocessing are provided in Section \ref{experiments}. For each input feature setting, we perform a search over 54 hyperparameter combinations of learning rate, weight decay and dropout.

\begin{table*}[h]
\centering
\caption{Mean Classification Accuracy on fully-supervised node classification task on 2-layered MLP with hidden dimension size as 64. \textbf{CAT} and \textbf{SUM} refers to concatenation and sum aggregation operation respectively.}
\label{tab:feat_imp}
\resizebox{\linewidth}{!}{%
\begin{tabular}{lcccccccccccl} 
\toprule
\multirow{2}{*}{\textbf{Dataset}} & \multicolumn{7}{c}{\begin{tabular}[c]{@{}c@{}}\textbf{Single\_feature}\\\end{tabular}}                                                                                        & \multicolumn{2}{c}{\textbf{All\_feature}} & \multicolumn{2}{c}{\begin{tabular}[c]{@{}c@{}}\textbf{Sub\_feature}\\\end{tabular}} & \multirow{2}{*}{\textcolor{denim}{\textbf{SOTA}}}  \\
                                  & $\mathbf{X}$ & $\mathbf{AX}$ & \textbf{$\mathbf{(A+I)X}$} & \textbf{$\mathbf{A^2X}$} & \textbf{$\mathbf{(A+I)^2X}$} & \textbf{$\mathbf{A^3X}$} & \textbf{$\mathbf{(A+I)^3X}$} & \textbf{CAT} & \textbf{SUM}               & \textbf{CAT}   & \textbf{SUM}                                                       &                \\ 
\hline
\textbf{Cora }                    & 73.40        & 79.55         & 84.28                      & 83.86                    & 85.47                        & 83.58                    & 85.41                        & 87.68        & 87.5                       & \textbf{88.10} & 88.04                                                              & \textcolor{denim}{88.49 \cite{chien_adaptive_2021,chen_simple_2020}}          \\
\textbf{Citeseer }                & 71.66        & 69.10         & 73.53                      & 72.38                    & 74.07                        & 70.55                    & 73.92                        & 77.08        & 77.09                      & \textbf{77.52} & 77.43                                                              & \textcolor{denim}{77.99 \cite{pei_geom-gcn_2020}}          \\
\textbf{Pubmed }                  & 87.79        & 81.77         & 88.27                      & 84.70                    & 88.06                        & 83.06                    & 86.63                        & 89.75        & 89.55                      & \textbf{89.88} & 89.83                                                              & \textcolor{denim}{90.30 \cite{chen_simple_2020}}         \\
\textbf{Chameleon }               & 46.05        & 77.74         & 71.22                      & 76.07                    & 71.77                        & 75.26                    & 71.62                        & 75.61        & 72.25                      & \textbf{78.59} & 78.55                                                              & \textcolor{denim}{66.47 \cite{chien_adaptive_2021}}          \\
\textbf{Wisconsin }               & 87.45        & 63.13         & 58.03                      & 62.54                    & 52.94                        & 60.00                    & 51.76                        & 85.09        & 79.8                       & 87.84          & \textbf{88.62}                                                     & \textcolor{denim}{86.98 \cite{suresh_breaking_2021}}          \\
\textbf{Texas }                   & 85.40        & 66.21         & 61.35                      & 67.29                    & 58.64                        & 62.43                    & 58.10                        & 84.32        & 78.91                      & 88.64          & \textbf{88.91}                                                     & \textcolor{denim}{86.49 \cite{chien_adaptive_2021}}          \\
\textbf{Cornell }                 & 85.94        & 58.64         & 63.51                      & 58.64                    & 61.62                        & 58.91                    & 60.27                        & 81.89        & 72.25                      & 86.21          & \textbf{86.75}                                                     & \textcolor{denim}{82.16 \cite{zhu_beyond_2020}}          \\
\textbf{Squirrel }                & 30.24        & 73.18         & 63.79                      & 71.28                    & 63.37                        & 64.42                    & 62.82                        & 73.02        & 64.68                      & \textbf{74.16} & 73.12                                                              & \textcolor{denim}{49.03 \cite{chien_adaptive_2021}}          \\
\textbf{Actor }                   & 35.32        & 25.47         & 29.22                      & 25.38                    & 27.95                        & 25.27                    & 26.43                        & 35.15        & 35.39                      & 35.63          & \textbf{35.67}                                                     & \textcolor{denim}{36.53 \cite{suresh_breaking_2021}}          \\
\bottomrule
\end{tabular}
}
\end{table*}

\subsection{Analysis of results}
\label{sec:analysis}
Table \ref{tab:feat_imp} shows the mean node classification accuracy with different input features for hidden dimension size of 64. We also include current state-of-the-art (SOTA) results for a comparison with our results. We find many interesting observations as follows:

\noindent
\textbf{Observation 1.} We find that each hop features contribute differently to the prediction performance of the model. Some hop features are more informative than the others. For homophily datasets: Cora, Citeseer, and Pubmed self-looped features have higher node classification accuracy. In many recent publications that have considered heterophily, there is often more emphasis on Texas, Wisconsin and Cornell as good heterophily datasets, and Squirrel and Chameleon are considered to have low-quality node features as heterophily datasets \cite{zhu_beyond_2020}. However, we observe that for Wisconsin, Texas, Cornell and Actor, the best features are node's own features, and features of the neighbors are not informative enough. In case of Squirrel and Chameleon, we achieve best performance with node's first hop no-loop features. In these two datasets, node's own features and self-looped features have a low correlation with node's labels. Hence, they are, in fact, very good representation of heterophily datasets. These observations highlight the importance of using both self-looped and no-looped adjacency matrices in GNNs for better generalization over homophily and heterophily datasets respectively.

Based on our above observations, we postulate that \textit{the problem of homophily and heterophily in graphs is a feature generation problem}. With appropriate feature generation measures, GNNs can learn on different types of graph datasets.

\noindent
\textbf{Observation 2.} In \texttt{All\_feature} setting, we train the model on all features with both concatenation and sum as aggregation operation. Our first observation is improved performance compared to \texttt{Single\_feature}, which is natural as all features in combination provide more information. Comparing the aggregator schemes, for many datasets: Chameleon, Wisconsin, Texas, Cornell, and Squirrel sum operation has significantly lower accuracy compared to concatenation operation even with higher dimension embeddings (Please refer to Table \ref{tab:feat_imp_high} for additional results with d=128 \& 256). In this setting, we find concatenation operation overall provides better accuracy values compared with sum operation.

\noindent
\textbf{Observation 3.} In \texttt{Sub\_feature} setting, we find significant improvements in performance of the model on all datasets compared to both \texttt{Single\_feature} setting and \texttt{All\_feature} settings. This observation implies that among all features, there are subsets of features that are more informative than others for better prediction performance. In addition, other less informative features, if present in the input, can act as noise and may lead to worse performance of the model. This leads to the idea that feature selection is an important aspect of the design of graph neural networks. By reducing the effect of less informative/noisy features, higher prediction accuracy can be achieved even with a simple two-layered neural network. Over-smoothing problem in GNNs is considered to be due to node features becoming less informative because of feature averaging over many hops. However, with a feature selection mechanism, these uninformative features can be ignored. With these observations, we postulate that \textit{graph learning and over-smoothing mitigation is a feature selection problem.} With a good feature selection strategy, GNN model can provide good prediction accuracy and eliminate the over-smoothing problem.

% \textcolor{blue}{Comparing with SOTA results, we find for homophily datasets, \texttt{Sub\_feature} has comparable results compared to more complex GNN models with difference in results decreasing with increase in embedding dimensions (Table \ref{tab:feat_imp_high}). For homo}

In addition, we find another interesting observation that in \texttt{Sub\_feature} setting, the difference in the performance of concatenation and sum aggregation operation is reduced and is not as significant as observed with \texttt{All\_feature} setting.

\begin{figure*}[h]
    \centering
    \includegraphics[width=0.68\textwidth]{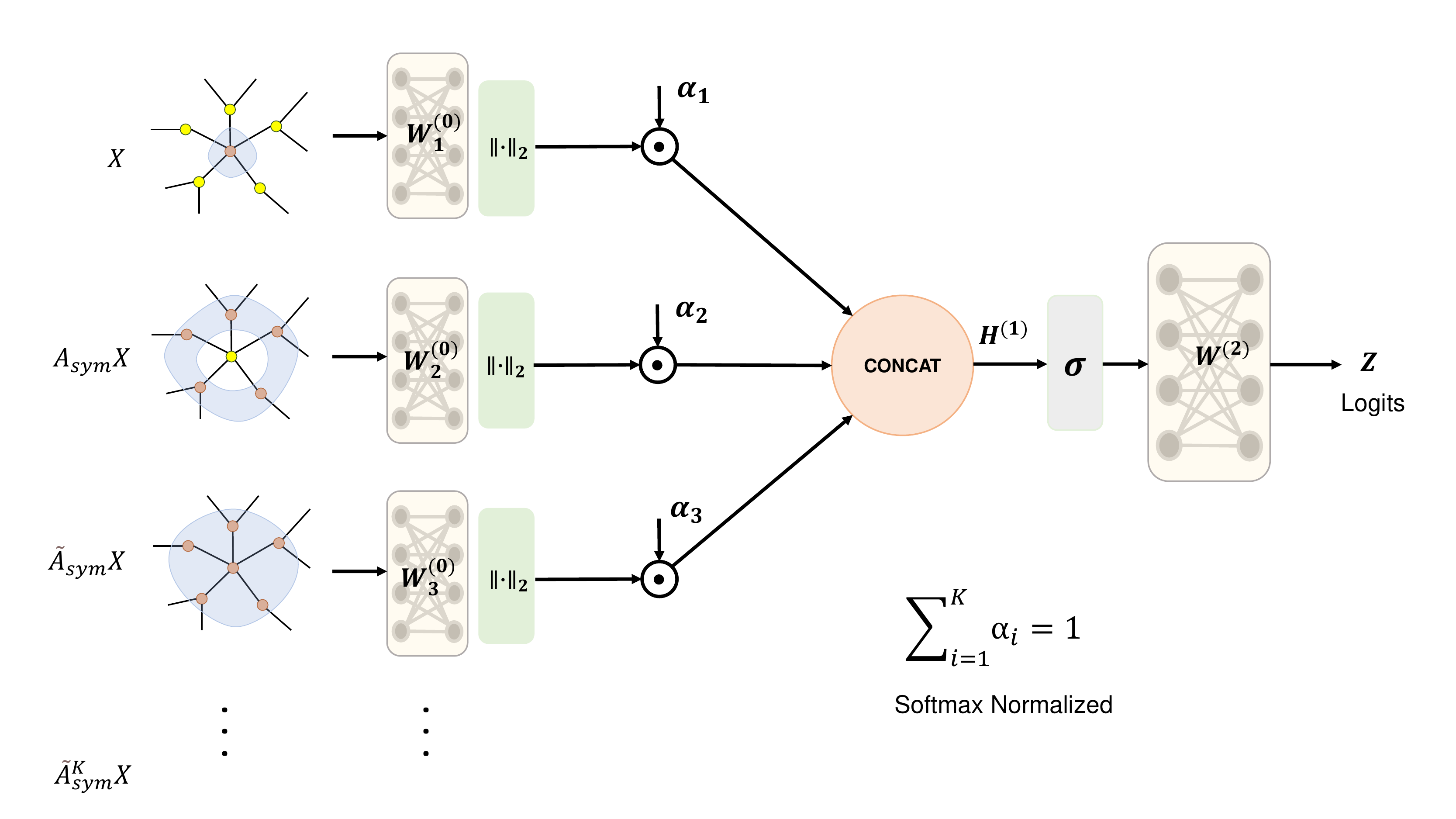}
    \caption{Figure shows model diagram of FSGNN. Input features are generated based on powers of $A$ and $\Tilde{A}$.}
    \label{fig:model_diagram}

\end{figure*}

\section{Proposed Architecture}
\label{propose_arch}

As discussed in Section \ref{sec:analysis} that feature selection is important to improve prediction capability of the model. However, using \texttt{Sub\_Feature} is not feasible for real-world applications as the number of input feature combinations increase exponentially with an increase in the number of hops, making it computationally expensive. Nevertheless, a GNN model can be designed to approximate feature selection strategy. When all input features are provided, the model should be able learn to assign higher weights to more relevant and informative features and actively reject features that are not useful. To construct such a model, we propose to weight input features with a single scalar value that is multiplied to each input feature matrix. We impose a constraint on these scalar values by the softmax function as follows. Let $\alpha_i$ be the scalar value for the $i^{th}$ feature matrix, then $\alpha_i$ scales the magnitude of the features as $\alpha_i X_iW^{(0)}_i$. Softmax function is used in deep learning as a non-linear normalizer, and its output is often practically interpreted as probabilities. Before training, the scalar values corresponding to each feature matrix are initialized with equal values, and softmax is applied on these values. The resultant normalized values $\alpha_i$ are then multiplied with the input features, and the concatenation operator is applied. Considering $L$ number of input feature matrices $X_l,\: l\in\{1\:..\:L\}$ , the formulation can be described as,

\begin{equation}
    %H^{(2)} = \textrm{CAT}(\alpha_1W_{1}^{(0)}X_l,\alpha_2W_{2}^{(0)}X_2, ... , \alpha_lW_{L}^{(0)}X_L)
    %H^{(2)} = CAT(\alpha_1W_{l}^{(0)}X_l)
    H^{(1)} = \bigparallel^{L}_{l=1} \alpha_lX_lW_{l}^{(0)}
\end{equation}
$ \textrm{where         }   \sum_{l=1}^{L}\alpha_{l} = 1$ and $\parallel$ denotes concatenation operation.
\newline

 While training, the scalar values of relevant features corresponding to the labels increase towards 1 while others decrease towards 0. The features that are not useful and represent more noise than signal have their magnitudes reduced with a corresponding decrease in their scalar values. Since we are not using a binary selection of features, we term this selection procedure as "soft-selection" of features.

The formulation discussed above can be understood in two ways. As GNNs have been represented with a polynomial filter, 

\begin{equation}
    g_\theta(P) = \sum_{k=0}^{K-1}\theta_kP^k
\end{equation}

\noindent
where $\theta \in \mathbb{R}^K$ is a vector of polynomial coefficients and P can be adjacency matrix \cite{kipf_semi-supervised_2017}\cite{chen_simple_2020}\cite{chien_adaptive_2021}, laplacian matrix \cite{nt_stacked_2020} or PageRank based matrix \cite{berberidis_adaptive_2019}. As the polynomial coefficients are scalar parameters then our scheme can be considered as applying regularization on these parameters using the softmax function. The other way to look is to simply consider it as a weighting scheme. The input features can be arbitrarily chosen, and instead of a scalar weighting scheme, a more sophisticated scheme can be used. 

 For practical implementation since all weights are initialized as equal, they are all set to 1. After normalizing with softmax function, the individual scalar values become equal to $1/L$. During training, these values change, denoting the importance of the features. As the scalar values affect the magnitude of the features, they also affect the gradients propagated back to the linear layer, which transforms the input features. Hence it is important to have a unique weight matrix for each input feature matrix.

%  In some cases, initial $\alpha_l = 1/L$ value may be too small and may adversely affect training. In that case, a constant $\gamma$ may be multiplied after softmax normalization to increase the initial magnitude as $\gamma\alpha_lX_lW_{l}^{(0)}$. Since $\gamma$ remains constant during the training, it does not affect the softmax regularization of the scalar parameters. 

\subsection*{Feature Selection Graph Neural Network}

Combining the model designs formulated earlier, we propose a simple and shallow (2-layered) graph GNN model called Feature Selection Graph Neural Network (FSGNN). Figure \ref{fig:model_diagram} shows the diagrammatic representation of our model. Input features are precomputed using $A_{sym}$ and $\Tilde{A}_{sym}$ and transformed using a linear layer unique to each feature matrix. L2-normalization is applied on the output activations of the first layer and weighted with scalar weights regularized by the softmax function. Output features are then concatenated and non-linearly transformed using ReLU and mapped to the second linear layer. Cross-entropy loss is calculated with output logits of the second layer. The model can be represented as,

\begin{equation}
    Z = \sigma \left( \mathrm{CONCAT}_l(\alpha_l X_l W^{(0)}_l) \right)W^{(1)}
\end{equation}

\noindent
where $\mathrm{CONCAT}_l, \forall \, l \in \{1\:..\:L\}$, $X_l$ are input features and $\sigma$ is ReLU activation function.

% \setlength{\textfloatsep}{1mm}
% \begin{algorithm}[h!]
% %\SetAlgoLined
% \DontPrintSemicolon
% \caption{Pseudo Code FSGNN (Forward propagation)}
% \label{alg:fsgnn}
%     \SetKwInOut{Input}{Input}\SetKwInOut{Output}{Output}
%     \Input{~ $A_{sym}$; $\Tilde{A}_{sym}$; No. of hops $K$; weight matrices $W^{(k)}$; $\alpha$ vector of dimension 2K+1;   }
%     \Output{~ Logits}
%     \BlankLine
%     \SetNoFillComment
%     \tcc{Precompute input features}
%     \For{$k=1...K$}{
%         $X_A \leftarrow A_{sym}X_A$ \;
%         $X_{\Tilde{A}} \leftarrow \Tilde{A}_{sym}X_{\Tilde{A}}$ \;
%         $list\_mat.APPEND(\:X_A\:)$ \;
%         $list\_mat.APPEND(\:X_{\Tilde{A}}\:)$ \;
%         }
%     \tcc{Model training}
%     $\alpha_i\leftarrow1.0,  i=1...2K+1$ \;
%     $\alpha \leftarrow SOFTMAX(\alpha)$ \;
%     $list\_mat \leftarrow [X]$ \;
%     $X_A\leftarrow X$ \; 
%     $X_{\Tilde{A}} \leftarrow X$ \;

%     $list\_cat = LIST()    $\;
%     \For{$j=1...2K+1$}{
%         $X_f \leftarrow list\_mat[j]$ \;
%         $Out \leftarrow L2\_NORM(\:X_fW_{j}^{(0)}\:) $ \;
%         $list\_cat.APPEND(\:\alpha_j \odot Out\:)$ \;
%     }
%     $H^{(1)} \leftarrow CONCAT(\:list\_cat\:)$ \;
%     $Z \leftarrow ReLU(\: H^{(1)}\:)W^{(2)}$

% \end{algorithm}

\section{Related Work}
\label{related_work}

GNNs have emerged as an indispensable tool to learn graph-centric data. Many prediction tasks like node classification, link prediction, graph classification, etc. \cite{defferrard_convolutional_2016}\cite{kipf_semi-supervised_2017} introduced a simple end-to-end training framework using approximations of spectral graph convolutions. Since then, there has been efforts in the research community to improve the performance of GNNs, and a variety of techniques have been introduced. Earlier GNN frameworks utilized a fixed propagation scheme along all edges, which is not always scalable for larger graphs. GraphSAGE\cite{hamilton_inductive_2017} and FastGCN\cite{chen_fastgcn_2018} introduce neighbor sampling approaches in graph neural networks. GAT \cite{velickovic_graph_2018} introduces the use of the attention mechanism to provide weights to features that are aggregated from the neighbors. APPNP \cite{klicpera_predict_2018}, JK \cite{xu_representation_2018}, Geom-GCN \cite{pei_geom-gcn_2020}, SimP-GCN \cite{jin_node_2021}, and CPGNN \cite{zhu_graph_2021} aim to improve the feature propagation scheme within layers of the model. More recently, researchers are proposing to make GNN models deeper \cite{chen_simple_2020,li_training_2021,godwin_very_2021}. However, deeper models suffer from over-smoothing, where after stacking many GNN layers, features of the node become indistinguishable from each other, and there is a drop in the performance of the model. DropEdge \cite{rong_dropedge_2020} proposes to drop a certain number of edges to reduce the speed of convergence of over-smoothing and relieves the information loss. GCNII \cite{chen_simple_2020} use residual connections and identity mapping in GNN layers to enable deeper networks. RevGNN \cite{li_training_2021} uses deep reversible architectures and \cite{godwin_very_2021} uses noise regularisation to train deep GNN models. 

Researchers find that traditional GNNs work well in homophily graphs but fail to generalize to heterophily graphs. Several models that are explicitly designed to handle heterophily graphs are proposed, including H2GCN \cite{zhu_beyond_2020}, CPGNN \cite{zhu_graph_2021}, TDGNN \cite{wang_tree_2021}, Geom-GCN \cite{pei_geom-gcn_2020}, and GPRGNN \cite{chien_adaptive_2021}. However, a recent work \cite{ma_is_2021} reveals that GCNs can achieve strong performance on heterophily graphs under certain conditions.

% \liu{Similar idea of decoupling feature generation and representation learning has been adopted by several existing works, such as SGC \cite{wu_simplifying_2019} and SIGN \cite{frasca_sign_2020}. However, they aim at increasing the model scalability by precomputing the features, whereas FSGNN targets at enhancing the model performance by selecting critical features.}

Similar to our work, the idea of decoupling feature generation and representation learning has been adopted by several existing works, such as SGC \cite{wu_simplifying_2019} and SIGN \cite{frasca_sign_2020}. However, these models are equivalent to \texttt{Single\_feature} and \texttt{All\_feature} setting, thus suffer from similar drawbacks. Many GNN models exhibit similarity of weighting the features, however, many of them have fixed weighting scheme like JK-Net \cite{xu_representation_2018}, APPNP \cite{klicpera_predict_2018}, and GCNII \cite{chen_simple_2020}. In case of models with adaptable weighting scheme like ChebyNet \cite{defferrard_convolutional_2016} and GPR-GNN \cite{chien_adaptive_2021}, learned weights do not show feature selection pattern due to lack of explicit regularization. Moreover, these GNN models do not use no-loop features by design, thus limiting their learning capability on heterophily datasets.

\section{Experiments}
\label{experiments}
In this section, we evaluate the empirical performance of our proposed model on real-world datasets on the node classification task
and compare with other graph neural network models.

\subsection{Datasets}

For fully-supervised node classification tasks, we perform experiments on nine datasets commonly used in graph neural networks literature. Details of the datasets are presented in Table \ref{tab:fully_supervised_data}. Homophily ratio \cite{zhu_beyond_2020} denotes the fraction of edges which connects two nodes of the same label. A higher value (closer to 1) indicates strong homophily, while a lower value (closer to 0) indicates strong heterophily in the dataset. Cora, Citeseer, and Pubmed \cite{sen_collective_2008} are citation networks based datasets and in general, are considered as homophily datasets. Graphs in Wisconsin, Cornell, Texas \cite{pei_geom-gcn_2020} represent links between webpages, Actor \cite{tang_social_2009} represent actor co-occurrence in Wikipedia pages, Chameleon and Squirrel \cite{rozemberczki_multi-scale_2020} represent the web pages in Wikipedia discussing corresponding topics. These datasets are considered as heterophily datasets. To provide a fair comparison, we use publicly available data splits taken from \cite{pei_geom-gcn_2020}\footnote{https://github.com/graphdml-uiuc-jlu/geom-gcn}. These splits have been frequently used by researchers for experiments in their publications. The results of comparison methods presented in this paper are also based on this split.

\begin{table}
\centering
\caption{Statistics of the node classification datasets}
\label{tab:fully_supervised_data}
\resizebox{\linewidth}{!}{%
\begin{tabular}{lcrrrc} 
\hline
\multicolumn{1}{c}{\textbf{Datasets}} & \textbf{Hom. Ratio} & \textbf{Nodes} & \textbf{Edges} & \multicolumn{1}{c}{\textbf{Features}} & \textbf{Classes}  \\ 
\hline
Cora                                  & 0.81                & 2,708          & 5,429          & 1,433                                 & 7                 \\
Citeseer                              & 0.74                & 3,327          & 4,732          & 3,703                                 & 6                 \\
Pubmed                                & 0.80                & 19,717         & 44,338         & 500                                   & 3                 \\
Chameleon                             & 0.23                & 2,277          & 36,101         & 2,325                                 & 4                 \\
Wisconsin                             & 0.21                & 251            & 499            & 1,703                                 & 5                 \\
Texas                                 & 0.11                & 183            & 309            & 1,703                                 & 5                 \\
Cornell                               & 0.30                & 183            & 295            & 1,703                                 & 5                 \\
Squirrel                              & 0.22                & 5,201          & 198,353        & 2,089                                 & 5                 \\
Actor                                 & 0.22                & 7,600          & 26,659         & 932                                   & 5                 \\
\hline
\end{tabular}
}
\end{table}

\subsection{Preprocessing}

We follow the same preprocessing steps used by \cite{pei_geom-gcn_2020} and \cite{chen_simple_2020}. Other models to which we compare our results also follow the same set of procedures. Initial node features are row-normalized. To account for both homophily and heterophily, we use the adjacency matrix and adjacency matrix with added-self loops for feature transformation. Both matrices are symmetrically normalized. For efficient computation, adjacency matrices are stored and used as sparse matrices.

\begin{table*}
\centering
\caption{Mean classification accuracy on fully-supervised node classification task. Results for GCN, GAT, GraphSAGE, Cheby+JK, MixHop and H2GCN-1 are taken from \cite{zhu_beyond_2020}. For GEOM-GCN, GCNII and WRGAT results are taken from the respective article. Best performance for each dataset is marked as bold and second best performance is underlined for comparison. }
\label{tab:full_super_results}
\resizebox{\linewidth}{!}{%
\begin{tabular}{lccccccccccc} 
\toprule
\multicolumn{2}{l}{}                                          & \textbf{Cora}           & \textbf{Citeseer}      & \textbf{Pubmed}         & \textbf{Chameleon}      & \textbf{Wisconsin}      & \textbf{Texas}          & \textbf{Cornell}        & \textbf{Squirrel}       & \textbf{Actor}          & \textbf{Mean Acc.}      \\ 
\hline
\multicolumn{2}{l}{\textbf{GCN}}                              & 87.28$\pm$1.26          & 76.68$\pm$1.64         & 87.38$\pm$0.66          & 59.82$\pm$2.58          & 59.80$\pm$6.99          & 59.46$\pm$5.25          & 57.03$\pm$4.67          & 36.89$\pm$1.34          & 30.26$\pm$0.79          & 61.62                   \\
\multicolumn{2}{l}{\textbf{GAT}}                              & 82.68$\pm$1.80          & 75.46$\pm$1.72         & 84.68$\pm$0.44          & 54.69$\pm$1.95          & 55.29$\pm$8.71          & 58.38$\pm$4.45          & 58.92$\pm$3.32          & 30.62$\pm$2.11          & 26.28$\pm$1.73          & 58.55                   \\
\multicolumn{2}{l}{\textbf{GraphSAGE}}                        & 86.90$\pm$1.04          & 76.04$\pm$1.30         & 88.45$\pm$0.50          & 58.73$\pm$1.68          & 81.18$\pm$5.56          & 82.43$\pm$6.14          & 75.95$\pm$5.01          & 41.61$\pm$0.74          & 34.23$\pm$0.99          & 69.50                   \\
\multicolumn{2}{l}{\textbf{Cheby+JK}}                         & 85.49$\pm$1.27          & 74.98$\pm$1.18         & 89.07$\pm$0.30          & 63.79$\pm$2.27          & 82.55$\pm$4.57          & 78.38$\pm$6.37          & 74.59$\pm$7.87          & 45.03$\pm$1.73          & 35.14$\pm$1.37          & 69.89                   \\
\multicolumn{2}{l}{\textbf{MixHop}}                           & 87.61$\pm$0.85          & 76.26$\pm$1.33         & 85.31$\pm$0.61          & 60.50$\pm$2.53          & 75.88$\pm$4.90          & 77.84$\pm$7.73          & 73.51$\pm$6.34          & 43.80$\pm$1.48          & 32.22$\pm$2.34          & 68.10                   \\
\multicolumn{2}{l}{\textbf{GEOM-GCN}}                         & 85.27                   & \textbf{77.99}         & \uline{90.05}           & 60.90                   & 64.12                   & 67.57                   & 60.81                   & 38.14                   & 31.63                   & 64.05                   \\
\multicolumn{2}{l}{\textbf{GCNII}}                            & 88.01$\pm$1.33          & 77.13$\pm$1.38         & \textbf{90.30$\pm$0.37} & 62.48$\pm$2.74          & 81.57$\pm$4.98          & 77.84$\pm$5.64          & 76.49$\pm$4.37          & N/A                     & N/A                     & -                       \\
\multicolumn{2}{l}{\textbf{H2GCN-1}}                          & 86.92$\pm$1.37          & 77.07$\pm$1.64         & 89.40$\pm$0.34          & 57.11$\pm$1.58          & 86.67$\pm$4.69          & 84.86$\pm$6.77          & 82.16$\pm$4.80          & 36.42$\pm$1.89          & \uline{35.86$\pm$1.03}  & 70.71                   \\
\multicolumn{2}{l}{\textbf{WRGAT}}                            & 88.20$\pm$2.26          & 76.81$\pm$1.89         & 88.52$\pm$0.92          & 65.24$\pm$0.87          & 86.98$\pm$3.78          & 83.62$\pm$5.50          & 81.62$\pm$3.90          & 48.85$\pm$0.78          & \textbf{36.53$\pm$0.77} & 72.93                   \\
\multicolumn{2}{l}{\textbf{GPRGNN}}                           & \textbf{88.49$\pm$0.95} & 77.08$\pm$1.63         & 88.99$\pm$0.40          & 66.47$\pm$2.47          & 85.88$\pm$3.70          & 86.49$\pm$4.83  & 81.89$\pm$6.17          & 49.03$\pm$1.28          & 36.04$\pm$0.96          & 73.37                   \\ 
\hline
\multirow{2}{*}{\textbf{FSGNN (Homo/Hetero)}} & \textbf{3-hop} & 87.61$\pm$1.39          & 77.17$\pm$1.48         & 89.70$\pm$0.44          & \uline{78.93$\pm$1.03}  & \uline{88.24$\pm$3.40}  & \textbf{87.57$\pm$4.71} & \uline{87.30$\pm$5.93}  & 73.86$\pm$1.81          & 35.38$\pm$0.81          & 78.42                   \\
                                             & \textbf{8-hop} & \uline{88.23$\pm$1.17}  & 77.35$\pm$1.17         & 89.78$\pm$0.38          & \textbf{78.95$\pm$0.86} & 87.65$\pm$3.51          & \textbf{87.57$\pm$4.86} & \uline{87.30$\pm$4.53}  & \uline{73.94$\pm$2.02}  & 35.62$\pm$0.87          & \textbf{78.49}          \\
\multirow{2}{*}{\textbf{FSGNN (All)}}         & \textbf{3-hop} & 87.73$\pm$1.36          & 77.19$\pm$1.35         & 89.73$\pm$0.39          & 78.14$\pm$1.25          & \textbf{88.43$\pm$3.22} & \uline{87.30$\pm$5.55}  & 87.03$\pm$5.77          & 73.48$\pm$2.13          & 35.67$\pm$0.69          & 78.30                   \\
                                             & \textbf{8-hop} & 87.93$\pm$1.00          & \uline{77.40$\pm$1.93} & 89.75$\pm$0.39          & 78.27$\pm$1.28          & 87.84$\pm$3.37          & \uline{87.30$\pm$5.28}  & \textbf{87.84$\pm$6.19} & \textbf{74.10$\pm$1.89} & 35.75$\pm$0.96          & \uline{78.46}  \\
\bottomrule
\end{tabular}
}
\end{table*}

\subsection{Settings and Baselines}

For a fully-supervised node classification task, each dataset is split evenly for each class into 48\%, 32\%, and 20\% for training, validation, and testing \cite{pei_geom-gcn_2020,zhu_beyond_2020}. We report the performance as mean classification accuracy over 10 random splits.

We fix the embedding size to 64, similar to other methods and set the initial learnable scalar parameter with respect to each hop to 1. Thus, the initial scalar value $\alpha_i$ is set to $1/L$. Hyper-parameter settings of the model for best performance are found by performing a grid-search over a range of hyper-parameters. We train the model under two input settings. In first setting, we follow the conventional classification of the datasets as homophily datasets and heterophily datasets. For homophily datasets, we use input features as node's self feature and self-looped aggregated features. For heterophily datasets, we use self-features and no-loop aggregated features. In the second setting, we use all features to train the model.
%Please refer to Appendix section for more details.

We compare our model to 10 different baselines and use the published results as the best performance of these models. GCNII \cite{chen_simple_2020} and H2GCN \cite{zhu_beyond_2020} have proposed multiple variants of their model. We have chosen the variant with the best performance on most datasets. GPRGNN uses random splits in their published results. For fair comparison, we ran their publicly available code on our standard splits while keeping other settings same. To get the best results, we performed hyperparameter search as mentioned in the repository. 

\section{Results}
\label{sec:results}

\subsection{Node Classification Results}
Table \ref{tab:full_super_results} shows the comparison of the mean classification accuracy of our model and other popular GNN models.
In general, traditional GNN models like GCN and GAT have higher performance on homophily datasets, however, they perform poorly on heterophily datasets. More recent models like H2GCN, WRGAT and GPRGNN perform relatively better on both homophily and heterophily datasets.   

On heterophily datasets, our model shows significant improvements especially 51.1\% on Squirrel and 18.8\% on Chameleon dataset. Similarly, on Wisconsin, Texas, and Cornell, improvements are 1.6\%, 1.2\%, and 6.9\%, respectively. On homophily datasets, we observe that different models perform best on different datasets. Our model still has consistent and comparable performance to SOTA.

\subsection{Comparison with \texttt{Sub\_feature}}

In our work, we aim to maintain performance as close as possible to \texttt{Sub\_feature} (hidden dimension, d = 64) in Table \ref{tab:feat_imp}. On five datasets: Cora, Chameleon, Cornell, Squirrel and Actor, our model performs as well as \texttt{Sub\_feature}, however for other datasets, performance is comparable, albeit a bit lower. During the training, our model starts with all input features and learns to identify relevant features and reduce the effect of irrelevant features. However, it is difficult to completely reduce the effect of noisy/irrelevant features without an explicit forgetting scheme. We consider the development of such a scheme to completely remove the impact of noisy features in GNN training as the future direction of our work.

\begin{table*}
\centering
\caption{Ablation study over 1080 different hyperparameter settings.}

\label{tab:ablation_study}
\resizebox{\linewidth}{!}{%
\begin{tabular}{lccccccccc} 
\toprule
                              & \textbf{Cora}           & \textbf{Citeseer}       & \textbf{Pubmed}         & \textbf{Chameleon}      & \textbf{Wisconsin}      & \textbf{Texas}          & \textbf{Cornell}        & \textbf{Squirrel}       & \textbf{Actor}           \\ 
\hline
\textbf{Proposed}             & 83.68$\pm$2.22          & 74.48$\pm$1.44          & \textbf{89.24$\pm$0.27} & \textbf{72.48$\pm$4.16} & 81.48$\pm$5.62          & \textbf{78.80$\pm$5.88} & \textbf{78.09$\pm$2.22} & \textbf{63.57$\pm$6.83} & 33.54$\pm$1.21           \\
\textbf{Without soft-selection} & \textbf{87.07$\pm$0.26} & \textbf{76.45$\pm$0.27} & 89.09$\pm$0.39          & 72.27$\pm$1.34          & 78.03$\pm$6.55          & 76.28$\pm$6.72          & 74.32$\pm$6.54          & 61.73$\pm$4.15          & 34.15$\pm$0.64           \\
\textbf{Common weight ($W^{(0)}$)}        & 83.19$\pm$1.41          & 72.15$\pm$1.02          & 88.96$\pm$0.28          & 68.24$\pm$6.03          & 70.56$\pm$10.94         & 68.45$\pm$7.65          & 68.18$\pm$9.13          & 56.63$\pm$8.54          & 32.73$\pm$1.48           \\
\textbf{Without L2-normalization}        & 77.12$\pm$3.49          & 71.40$\pm$10.01         & 87.72$\pm$0.77          & 53.06$\pm$6.18          & \textbf{82.60$\pm$2.68} & 76.33$\pm$3.87          & 76.18$\pm$3.43          & 32.60$\pm$6.38          & \textbf{36.66$\pm$0.55}  \\
\bottomrule
\end{tabular}
}
\end{table*}

\subsection{Ablation Studies}

In this section, we consider the effect of various proposed design strategies in section \ref{sec:design_strategy} on the performance of the model. In general, graph neural networks are sensitive to the hyperparameters used in training and require some amount of tuning to get the best performance. Since each dataset may have a different set of best hyperparameters, it can be difficult to judge design decisions based just on best performance of the model with single hyperparameter setting. To provide a comprehensive evaluation, we compare the average accuracy of the model over 1080 combinations of the hyperparameters. The hyperparameters we tune are learning rate and weight decay of layers and dropout value applied as regularization between layers.  Table \ref{tab:ablation_study} shows the average of classification accuracy values under various settings.

For most datasets, our proposed design schemes lead to better average accuracy. Cora and Citeseer show better average performance without softmax regularization; however, the peak performance is marginally less with regularization. Even though Wisconsin shows higher average accuracy without normalization, however, the best performance on the dataset was achieved with the normalization layer. We found that Actor was the only dataset where accuracy was  reduced with the addition of the normalization layer. Without the normalization layer, our model achieves 37.63\% accuracy. However, to maintain consistency, we do not include it in the main results. These variations also highlight that a single set of design choices may not apply to all datasets/tasks and some level of exploration is required.

It is interesting to note that performance on almost all datasets is sensitive to the choice of the hyperparameters for training the model as there is a wide gap between best and average performance. One exception is Pubmed, where the model's performance is relatively unperturbed under various hyperparameter combinations. 

\begin{figure}[h]
    \centering
    \includegraphics[width=\linewidth]{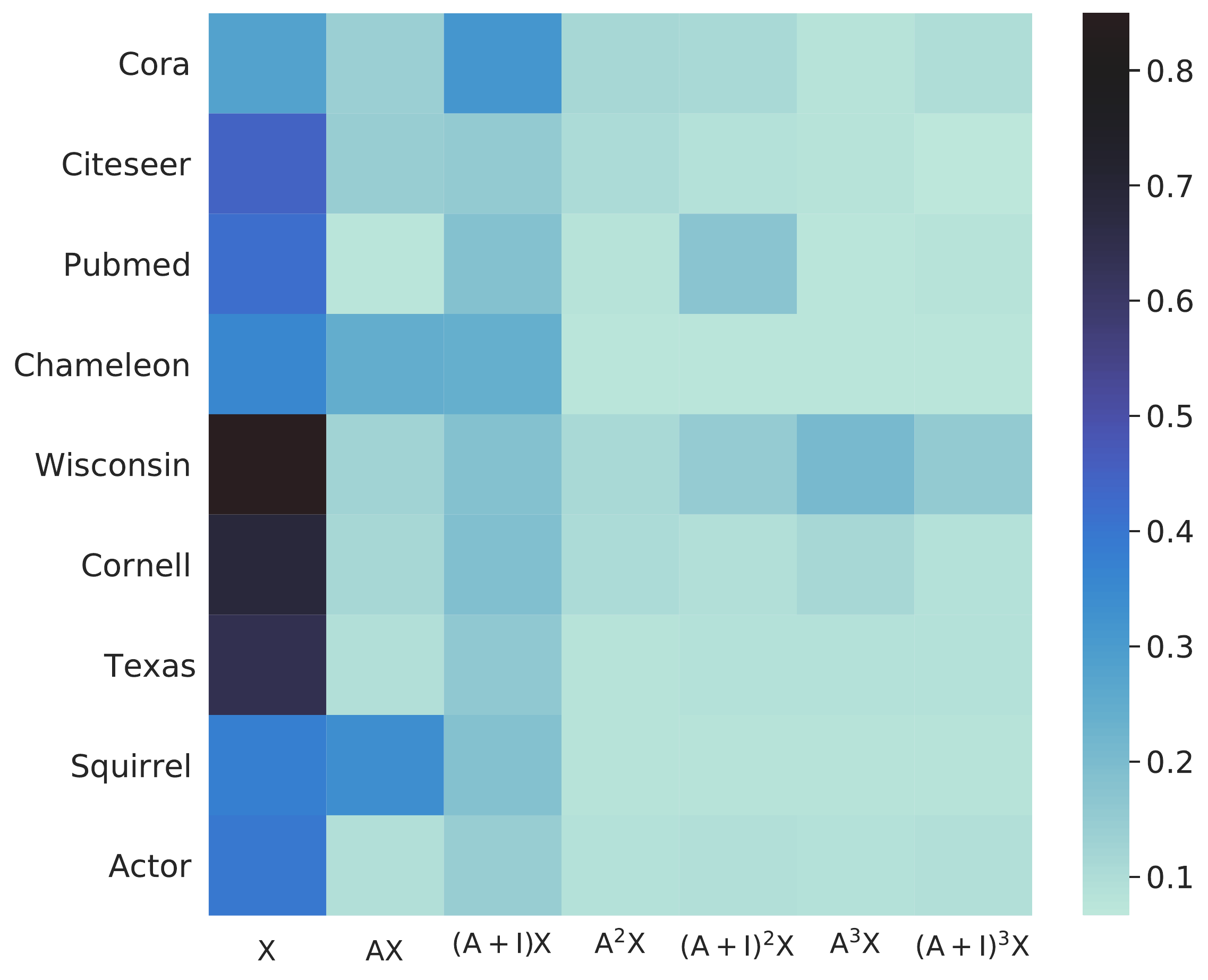}
    \caption{Heatmap of average of learned soft-selection scalar for all datasets}
    \label{fig:scalar_val_heat}
\end{figure}

\subsection{Soft-Selection Parameter Analysis}

We analyze the learned soft-selection parameters on average over different model hyperparameter combinations. We use four different settings: 1) Proposed model setting, 2) without softmax regularization on scalar weight parameters, 3) shared linear transformation layer on input features, and 4) without L2-normalization on input feature activations.  For homophily datasets, it is easy to see that self-looped features are given more importance. Among heterophily datasets, Wisconsin, Cornell, Texas, and Actor have the most weights on node's self features. In these datasets, graph structure plays a limited role in the performance accuracy of the model. For Chameleon and Squirrel datasets, we observed that the node's own features and first-hop features (without self-loop) were more useful for classification than any other features.

\begin{figure}[h]
    \centering
    \includegraphics[width=\linewidth]{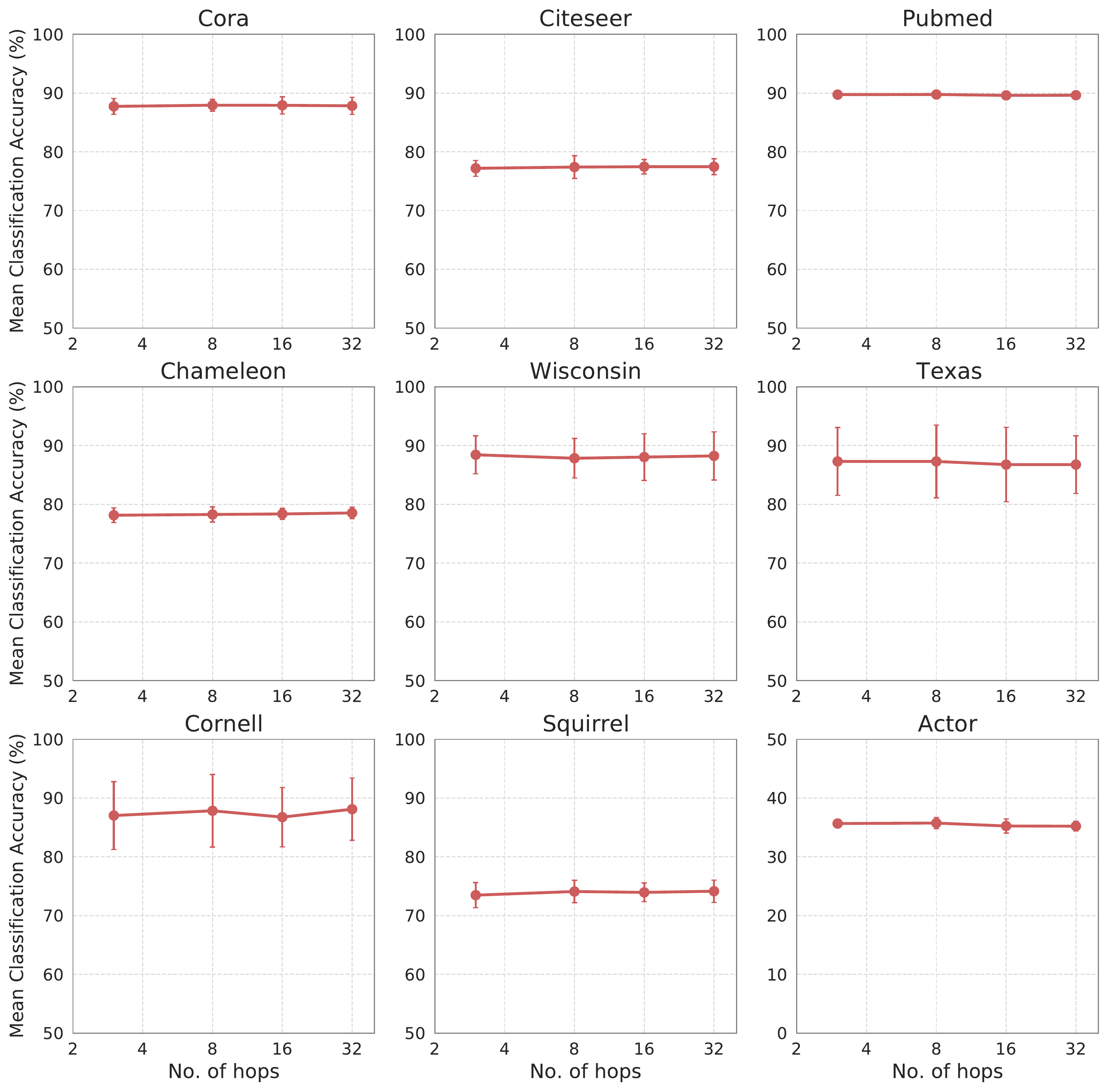}
    \caption{Figure shows the effect on classification accuracy of FSGNN with increase in the number of hops of feature aggregation. x-axis is in logarithmic scale. }
    \label{fig:accuracy_hop}
\end{figure}

\subsection{ Over-smoothing Analysis }

Many GNN models suffer from the over-smoothing problem when the number of hops for feature aggregation is increased. In section \ref{sec:analysis}, we discussed how feature selection can be helpful to overcome over-smoothing problem. In this section, we evaluate the change in model's performance with increase in the hops for aggregation. We run additional experiments with hop values set to 16, and 32 with all features as input as described in Section 6. Figure \ref{fig:accuracy_hop} shows the performance of the model for hop setting of 3,8,16 and 32. We observe that there is little variation in the performance of the model on various datasets and the model does not suffer from over-smoothing. This result is intuitive as aggregated features from higher hops are not very useful, and the model can learn to place low weights on them. For few datasets, we were able to achieve higher accuracy values: Citeseer (77.46\%), Cornell (88.11\%) and Squirrel (74.15\%).

%accuracy_hop.pdf (0.82)

\section{Conclusion}
\label{conclusion}

In this work, we explore the importance of feature selection in GNN training. We run extensive experiments to investigate GNN model design requirements for homophily and heterophily datasets and how feature selection can lead to higher prediction accuracy on benchmark datasets. Our experimental observations provide a definite confirmation that feature selection is a good direction for the exploration of GNN architectures. Based on our experimental observations, we propose a novel GNN model called FSGNN. Using extensive experiments, we show that FSGNN outperforms the current SOTA GNN models on the node classification task. Analysis of the learned parameters provides us the crucial information of feature importance. In addition, we show that even shallow models can learn and provide high prediction accuracy, and with our model over-smoothing phenomenon can be easily avoided. 

\section*{Acknowledgements}
This work was supported by JSPS Grant-in-Aid for Scientific Research (Grant Number 21K12042, 17H01785), JST CREST (Grant Number JPMJCR1687), and the New Energy and Industrial Technology Development Organization (Grant Number JPNP20006).

\clearpage
\appendix

%\newpage
\appendixpage

\section{Extended experiment results }

\begin{table*}
\centering
\caption{Mean Classification Accuracy on fully-supervised node classification task with hidden dimensions set to 64, 128, 256 \& 512.}
\label{tab:feat_imp_high}
\resizebox{\linewidth}{!}{%
\begin{tabular}{llcccccccccccc} 
\toprule
\multirow{2}{*}{\textbf{Dataset}}    & \multicolumn{1}{l}{\multirow{2}{*}{\textbf{\#dimensions}}} & \multicolumn{7}{c}{\begin{tabular}[c]{@{}c@{}}\textbf{Single\_Feature}\\\end{tabular}}                                           & \multicolumn{2}{c}{\textbf{All\_feature}} & \multicolumn{2}{c}{\begin{tabular}[c]{@{}c@{}}\textbf{Sub\_feature}\\\end{tabular}} & \multirow{2}{*}{\textbf{SOTA}}     \\
                                     & \multicolumn{1}{l}{}                                       & $\mathbf{X}$ & $\mathbf{AX}$ & $\mathbf{(A+I)X}$ & $\mathbf{A^2X}$ & $\mathbf{(A+I)^2X}$ & $\mathbf{A^3X}$ & $\mathbf{(A+I)^3X}$ & \textbf{CAT} & \textbf{SUM}               & \textbf{CAT}   & \textbf{SUM}                                                       &                                    \\ 
\hline
\multirow{4}{*}{\textbf{Cora }}      & d = 64                                                     & 73.40        & 79.55         & 84.28             & 83.86           & 85.47               & 83.58           & 85.41               & 87.68        & 87.5                       & 88.10          & 88.04                                                              & \multirow{4}{*}{\textcolor{denim}{88.49 \cite{chien_adaptive_2021,chen_simple_2020}}           }  \\
                                     & d = 128                                                      & 73.84        & 79.93         & 84.56             & 85.85           & 86.94               & 85.05           & 86.23               & 87.76        & 87.92                      & 88.19          & 88.43                                                              &                                    \\
                                     & d = 256                                                      & 74.06        & 82.25         & 85.97             & 86.27           & 87.54               & 85.67           & 86.86               & 87.70        & 87.68                      & 88.09          & 88.41                                                              &                                    \\
                                     & d = 512                                                      & 76.34        & 82.19         & 86.22             & 86.42           & 87.56               & 85.59           & 87.18               & 87.95        & 87.60                      & 87.97          & \textbf{88.53}                                                     &                                    \\ 
\hline
\multirow{4}{*}{\textbf{Citeseer }}  & d = 64                                                     & 71.66        & 69.10         & 73.53             & 72.38           & 74.07               & 70.55           & 73.92               & 77.08        & 77.09                      & 77.52          & 77.43                                                              & \multirow{4}{*}{\textcolor{denim}{77.99 \cite{pei_geom-gcn_2020}}           }  \\
                                     & d = 128                                                      & 71.94        & 70.74         & 73.96             & 73.70           & 74.58               & 71.42           & 74.28               & 77.35        & 77.04                      & 77.70          & 77.63                                                              &                                    \\
                                     & d = 256                                                      & 72.54        & 71.80         & 76.67             & 75.02           & 76.53               & 72.77           & 75.36               & 77.35        & 77.11                      & 77.86          & 77.74                                                              &                                    \\
                                     & d = 512                                                    & 73.57        & 71.95         & 76.62             & 75.11           & 76.83               & 73.19           & 75.91               & 77.32        & 77.35                      & 77.78          & \textbf{78.04}                                                     &                                    \\ 
\hline
\multirow{4}{*}{\textbf{Pubmed }}    & d = 64                                                     & 87.79        & 81.77         & 88.27             & 84.70           & 88.06               & 83.06           & 86.63               & 89.75        & 89.55                      & 89.88          & 89.83                                                              & \multirow{4}{*}{\textcolor{denim}{90.30 \cite{chen_simple_2020}}           }  \\
                                     & d = 128                                                      & 87.93        & 81.90         & 88.26             & 84.84           & 88.09               & 83.01           & 86.66               & 89.82        & 89.58                      & 89.92          & 89.86                                                              &                                    \\
                                     & d = 256                                                      & 88.01        & 81.89         & 88.31             & 84.86           & 88.08               & 83.02           & 86.74               & 89.81        & 89.64                      & \textbf{89.97} & 89.89                                                              &                                    \\
                                     & d = 512                                                    & 88.11        & 81.97         & 88.33             & 84.86           & 88.08               & 83.07           & 86.72               & 89.55        & 89.77                      & 89.90          & 89.88                                                              &                                    \\ 
\hline
\multirow{4}{*}{\textbf{Chameleon }} & d = 64                                                     & 46.05        & 77.74         & 71.22             & 76.07           & 71.77               & 75.26           & 71.62               & 75.61        & 72.25                      & 78.59          & 78.55                                                              & \multirow{4}{*}{\textcolor{denim}{66.47 \cite{chien_adaptive_2021}}           }  \\
                                     & d=128                                                      & 46.07        & 77.74         & 71.40             & 76.11           & 71.42               & 76.07           & 71.86               & 75.76        & 71.4                       & 78.99          & 77.98                                                              &                                    \\
                                     & d = 256                                                      & 46.09        & 77.63         & 71.25             & 76.77           & 71.07               & 76.2            & 72.58               & 76.77        & 70.81                      & 79.01          & 77.63                                                              &                                    \\
                                     & d = 512                                                      & 46.14        & 77.70         & 71.27             & 77.02           & 72.06               & 76.12           & 72.52               & 70.06        & 77.32                      & \textbf{79.07} & 77.30                                                              &                                    \\ 
\hline
\multirow{4}{*}{\textbf{Wisconsin }} & d = 64                                                     & 87.45        & 63.13         & 58.03             & 62.54           & 52.94               & 60.00           & 51.76               & 85.09        & 79.8                       & 87.84          & 88.62                                                              & \multirow{4}{*}{\textcolor{denim}{86.98 \cite{suresh_breaking_2021}}           }  \\
                                     & d = 128                                                      & 88.03        & 62.54         & 57.84             & 62.15           & 52.35               & 58.82           & 51.17               & 85.29        & 82.94                      & 88.43          & 88.04                                                              &                                    \\
                                     & d = 256                                                      & 88.03        & 62.54         & 58.03             & 61.96           & 51.76               & 57.84           & 50.78               & 87.45        & 83.92                      & 89.02          & 89.22                                                              &                                    \\
                                     & d = 512                                                      & 88.04        & 62.94         & 59.02             & 62.15           & 51.96               & 57.45           & 51.96               & 84.51        & 88.04                      & \textbf{89.41} & 90.0                                                               &                                    \\ 
\hline
\multirow{4}{*}{\textbf{Texas }}     & d = 64                                                     & 85.40        & 66.21         & 61.35             & 67.29           & 58.64               & 62.43           & 58.10               & 84.32        & 78.91                      & 88.64          & 88.91                                                              & \multirow{4}{*}{\textcolor{denim}{86.49 \cite{chien_adaptive_2021}}           }  \\
                                     & d=128                                                      & 86.21        & 67.02         & 61.62             & 67.56           & 58.64               & 61.62           & 57.83               & 84.32        & 78.91                      & 88.38          & 88.11                                                              &                                    \\
                                     & d = 256                                                      & 85.94        & 67.83         & 61.08             & 67.29           & 58.91               & 61.35           & 58.10               & 86.48        & 82.92                      & 88.65          & 88.65                                                              &                                    \\
                                     & d = 512                                                      & 85.67        & 67.03         & 61.89             & 67.30           & 59.19               & 61.35           & 58.65               & 83.51        & 86.49                      & 88.65          & \textbf{89.19}                                                     &                                    \\ 
\hline
\multirow{4}{*}{\textbf{Cornell }}   & d = 64                                                     & 85.94        & 58.64         & 63.51             & 58.64           & 61.62               & 58.91           & 60.27               & 81.89        & 72.25                      & 86.21          & 86.75                                                              & \multirow{4}{*}{\textcolor{denim}{82.16 \cite{zhu_beyond_2020}}           }  \\
                                     & d = 128                                                      & 86.21        & 58.10         & 63.78             & 58.64           & 60.54               & 58.91           & 60.27               & 84.05        & 74.86                      & 87.56          & 87.57                                                              &                                    \\
                                     & d = 256                                                      & 87.83        & 58.64         & 65.40             & 58.64           & 61.08               & 58.91           & 60.54               & 85.13        & 77.29                      & 88.11          & 87.57                                                              &                                    \\
                                     & d = 512                                                      & 87.30        & 59.19         & 65.13             & 58.92           & 62.16               & 58.92           & 60.81               & 81.89        & 86.76                      & \textbf{88.38} & 88.11                                                              &                                    \\ 
\hline
\multirow{4}{*}{\textbf{Squirrel }}  & d = 64                                                     & 30.24        & 73.18         & 63.79             & 71.28           & 63.37               & 64.42           & 62.82               & 73.02        & 64.68                      & 74.16          & 73.12                                                              & \multirow{4}{*}{\textcolor{denim}{49.03 \cite{chien_adaptive_2021}}           }  \\
                                     & d = 128                                                      & 30.30        & 72.83         & 63.66             & 71.49           & 64.43               & 64.49           & 63.59               & 72.55        & 62.50                      & 73.87          & 72.78                                                              &                                    \\
                                     & d = 256                                                      & 30.66        & 72.54         & 63.28             & 71.91           & 65.36               & 65.24           & 63.77               & 72.63        & 59.88                      & 74.49          & 72.76                                                              &                                    \\
                                     & d = 512                                                      & 30.74        & 73.11         & 63.22             & 72.24           & 65.84               & 65.48           & 64.38               & 58.02        & 72.87                      & \textbf{74.76} & 73.16                                                              &                                    \\ 
\hline
\multirow{4}{*}{\textbf{Actor }}     & d = 64                                                     & 35.32        & 25.47         & 29.22             & 25.38           & 27.95               & 25.27           & 26.43               & 35.15        & 35.39                      & 35.63          & 35.67                                                              & \multirow{4}{*}{\textcolor{denim}{36.53 \cite{suresh_breaking_2021}}           }  \\
                                     & d = 128                                                      & 35.75        & 25.38         & 29.26             & 25.25           & 27.71               & 25.26           & 26.21               & 35.94        & 35.57                      & 35.96          & 36.05                                                              &                                    \\
                                     & d = 256                                                      & 36.08        & 25.41         & 29.28             & 25.23           & 27.53               & 25.29           & 26.15               & 36.10        & 35.60                      & 36.22          & 36.31                                                              &                                    \\
                                     & d = 512                                                      & 36.38        & 25.38         & 29.26             & 25.42           & 27.39               & 25.26           & 26.08               & 36.02        & 36.34                      & \textbf{36.52} & 36.42                                                              &                                    \\
\bottomrule
\end{tabular}
}
\end{table*}

\subsection{Node classification on higher dimensions}
\label{sec:high_dimension}

Table \ref{tab:feat_imp_high} shows the mean node classification accuracy under \texttt{Single\_Feature}, \texttt{All\_Feature} and \texttt{Sub\_Feature} settings with hidden dimensions set to 64, 128, 256 and 512. We observe that just with increase in hidden dimensions of 2-layered MLP, we approach classification accuracy values similar or higher than state-of-the-art more complex and/or deeper GNN models. Thus the increase in number of parameters in the model help to improve the classification accuracy. However, there are diminishing returns over accuracy improvements with increasing the hidden dimensions of the model.

\subsection{Experiments with no non-linear activation}

As we observe in Table \ref{tab:feat_imp_high}, with \texttt{Sub\_Feature} scheme enabling feature selection, a simple 2-layered MLP already trains to very high accuracy. However, we would like to further understand the requirements of GNN complexity for the given benchmark datasets. In this section, we compare the effect of non-linear activation ReLU in 2-layered MLP model under \texttt{Sub\_Feature} setting with hidden dimensions set to 256. We remove only ReLU unit between the two layers and keep other settings same for learning rate, weight decay and dropout.

\begin{table}[h]
\centering
\caption{Mean node classification accuracy under \texttt{Sub\_Feature} setting with and without using ReLU activation and hidden dimensions set to 256.}
\label{tab:norelu_results}
\resizebox{0.9\linewidth}{!}{%
\begin{tabular}{lllllc} 
\toprule
\multirow{2}{*}{\textbf{Dataset}} & \multicolumn{2}{c}{\begin{tabular}[c]{@{}c@{}}\textbf{Sub\_Feature}\\\textbf{(With ReLU)}\end{tabular}} & \multicolumn{2}{c}{\begin{tabular}[c]{@{}c@{}}\textbf{Sub\_Feature}\\\textbf{(No ReLU)}\end{tabular}} & \multirow{2}{*}{\textcolor{denim}{\textbf{SOTA}}}  \\ 
\cline{2-5}
                                  & \textbf{CAT}   & \textbf{SUM}                                                                           & \textbf{CAT}   & \textbf{SUM}                                                                         &                                 \\ 
\hline
\textbf{Cora}                     & 88.09          & 88.41                                                                                  & 88.49          & \textbf{88.51}                                                                       & \textcolor{denim}{88.49}                           \\
\textbf{Citeseer}                 & \textbf{77.86} & 77.74                                                                                  & 77.72          & 77.81                                                                                & \textcolor{denim}{77.99}                           \\
\textbf{Pubmed}                   & \textbf{89.97} & 89.89                                                                                  & 89.52          & 89.54                                                                                & \textcolor{denim}{90.30}                           \\
\textbf{Chameleon}                & \textbf{79.01} & 77.63                                                                                  & 76.95          & 76.69                                                                                & \textcolor{denim}{66.47}                           \\
\textbf{Wisconsin}                & 89.02          & \textbf{89.22}                                                                         & 88.63          & 88.82                                                                                & \textcolor{denim}{86.98}                           \\
\textbf{Texas}                    & 88.65          & 88.65                                                                                  & \textbf{89.73} & 88.38                                                                                & \textcolor{denim}{86.49}                           \\
\textbf{Cornell}                  & 88.11          & 87.57                                                                                  & \textbf{88.38} & 87.84                                                                                & \textcolor{denim}{82.16}                           \\
\textbf{Squirrel}                 & \textbf{74.49} & 72.76                                                                                  & 70.45          & 69.94                                                                                & \textcolor{denim}{49.03}                           \\
\textbf{Actor}                    & 36.22          & \textbf{36.31}                                                                         & 36.11          & 35.8                                                                                 & \textcolor{denim}{36.53}                           \\
\bottomrule
\end{tabular}
}
\end{table}

Table \ref{tab:norelu_results} shows the accuracy comparisons of the models with and without non-linear activation between layers. Comparing both settings, we find two interesting observations. First, for Cora, Texas and Cornell datasets, we see further improvement in accuracy values. Second, for other datasets while accuracy values have decreased (expectedly), the difference is not significant except for Chameleon and Squirrel datasets.

With these results, we infer that for these datasets, simple graph convolution operation over node features combined with hop-feature selection provides sufficient information. Thus enabling a simple 2-layered model to perform well on the node classification task.

\section{Implementation Details of FSGNN}

For reproducibility of experimental results, we provide the details of our experiment setup and hyperparameters of the model. 

We use PyTorch 1.6.0 as deep learning framework on Python 3.8. Model training is done on Nvidia V100 GPU with 16 GB graphics memory and CUDA version 10.2.89.

For node classfication results (\ref{tab:full_super_results}), we do grid search for learning rate and weight decay of the layers and dropout between the layers. Hyperparameters are set for first layer $fc1$, second layer $fc2$ and scalar weight parameter $sca$. ReLU is used as non-linear activation and Adam is used as the optimizer. Table \ref{tab:param_search} shows details of hyperparameter search space. Table \ref{tab:3_hop_param} and \ref{tab:8_hop_param} show the best hyperparameters for the model in 3-hop and 8-hop configuration respectively. Patience value 100 is used for all datasets.

\begin{table}[h]
\centering
\caption{Hyperparameter search space for FSGNN. Experiments in Table \ref{tab:feat_imp} \& \ref{tab:feat_imp_high} do not use hyperparameters for scalar parameter.}
\label{tab:param_search}
\resizebox{0.85\linewidth}{!}{%
\begin{tabular}{ll} 
\toprule
\textbf{Hyperparameter} & \multicolumn{1}{c}{\textbf{Values}}  \\ 
\hline
\textbf{$WD_{sca}$}          & 0.0, 0.0001, 0.001, 0.01, 0.1        \\
\textbf{$LR_{sca}$}          & 0.04, 0.02, 0.01, 0.005              \\
\textbf{$WD_{fc1}$}          & 0.0, 0.0001, 0.001                   \\
\textbf{$WD_{fc2}$}          & 0.0, 0.0001, 0.001                   \\
\textbf{$LR_{fc}$}           & 0.01, 0.005                          \\
\textbf{$Dropout$}        & 0.5, 0.6, 0.7                        \\
\bottomrule
\end{tabular}
}
\end{table}

\begin{table}[h]
\centering
\caption{Hyperparameters of the 3-hop model (all-features) }
\label{tab:3_hop_param}
\resizebox{\linewidth}{!}{%
\begin{tabular}{lcccccc} 
\toprule
\multicolumn{1}{l}{\textbf{Datasets}} & \textbf{$WD_{sca}$} & \textbf{$LR_{sca}$} & \textbf{$WD_{fc1}$} & \textbf{$WD_{fc2}$} & \textbf{$LR_{fc}$} & \textbf{$Dropout$}  \\ 
\hline
\textbf{Cora}                         & 0.1            & 0.01           & 0.001          & 0.0001         & 0.01          & 0.6               \\
\textbf{Citeseer}                     & 0.0001         & 0.005          & 0.001          & 0.0            & 0.01          & 0.5               \\
\textbf{Pubmed}                       & 0.01           & 0.005          & 0.0001         & 0.0001         & 0.01          & 0.7               \\
\textbf{Chameleon}                    & 0.1            & 0.005          & 0.0            & 0.0            & 0.005         & 0.5               \\
\textbf{Wisconsin}                    & 0.0001         & 0.01           & 0.001          & 0.0001         & 0.01          & 0.5               \\
\textbf{Texas}                        & 0.001          & 0.01           & 0.001          & 0.0            & 0.01          & 0.7               \\
\textbf{Cornell}                      & 0.0            & 0.01           & 0.001          & 0.001          & 0.01          & 0.5               \\
\textbf{Squirrel}                     & 0.1            & 0.04           & 0.0            & 0.001          & 0.01          & 0.7               \\
\textbf{Actor}                        & 0.0            & 0.04           & 0.001          & 0.0001         & 0.01          & 0.7               \\
\bottomrule
\end{tabular}
}
\end{table}

\begin{table}[h]
\centering
\caption{Hyperparameters of the 8-hop model (all-features)}
\label{tab:8_hop_param}
\resizebox{\linewidth}{!}{%
\begin{tabular}{lcccccc} 
\toprule
\multicolumn{1}{l}{ \textbf{Datasets} } & \textbf{$WD_{sca}$}  & \textbf{$LR_{sca}$}  & \textbf{$WD_{fc1}$}  & \textbf{$WD_{fc2}$}  & \textbf{$LR_{fc}$}  & \textbf{$Dropout$}   \\ 
\hline
\textbf{Cora}                           & 0.1                  & 0.02                 & 0.001                & 0.0001               & 0.01                & 0.6                  \\
\textbf{Citeseer}                       & 0.0001               & 0.01                 & 0.001                & 0.0001               & 0.01                & 0.5                  \\
\textbf{Pubmed}                         & 0.01                 & 0.02                 & 0.0001               & 0.0                  & 0.005               & 0.7                  \\
\textbf{Chameleon}                      & 0.1                  & 0.01                 & 0.0                  & 0.0                  & 0.005               & 0.5                  \\
\textbf{Wisconsin}                      & 0.001                & 0.02                 & 0.001                & 0.0001               & 0.01                & 0.5                  \\
\textbf{Texas}                          & 0.01                 & 0.01                 & 0.001                & 0.0                  & 0.01                & 0.7                  \\
\textbf{Cornell}                        & 0.0                  & 0.01                 & 0.001                & 0.0001               & 0.01                & 0.5                  \\
\textbf{Squirrel}                       & 0.1                  & 0.02                 & 0.0                  & 0.0001               & 0.01                & 0.5                  \\
\textbf{Actor}                          & 0.0001               & 0.04                 & 0.001                & 0.0001               & 0.01                & 0.7                  \\
\bottomrule
\end{tabular}
}
\end{table}

% \begin{table}[h]
% \centering
% \caption{Hyperparameters for the ogbn-paper100M dataset}
% \label{tab:ogbn_papers}
% \resizebox{\linewidth}{!}{%
% \begin{tabular}{lccccccc} 
% \toprule
%  \textbf{Dataset}                                                   & \textbf{$WD_{sca}$}  & \textbf{$LR_{sca}$}  & \textbf{$WD_{fc1}$}  & \textbf{$WD_{fc2}$}  & \textbf{$LR_{fc1}$}& \textbf{$LR_{fc2}$}  & \textbf{$Dropout$}   \\ 
% \hline
% \begin{tabular}[c]{@{}l@{}}\textbf{ogbn-papers100M}\\ \end{tabular} & 0.1                  & 0.0001               & 0.001                & 0.000001             & 0.00005             & 0.0002              & 0.5                  \\
% \bottomrule
% \end{tabular}
% }
% \end{table}

\printcredits

\newpage
\bibliographystyle{model1-num-names}
%\bibliographystyle{cas-model2-names}

\bibliography{references}

\begin{thebibliography}{43}
\expandafter\ifx\csname natexlab\endcsname\relax\def\natexlab#1{#1}\fi
\providecommand{\url}[1]{\texttt{#1}}
\providecommand{\href}[2]{#2}
\providecommand{\path}[1]{#1}
\providecommand{\DOIprefix}{doi:}
\providecommand{\ArXivprefix}{arXiv:}
\providecommand{\URLprefix}{URL: }
\providecommand{\Pubmedprefix}{pmid:}
\providecommand{\doi}[1]{\href{http://dx.doi.org/#1}{\path{#1}}}
\providecommand{\Pubmed}[1]{\href{pmid:#1}{\path{#1}}}
\providecommand{\bibinfo}[2]{#2}
\ifx\xfnm\relax \def\xfnm[#1]{\unskip,\space#1}\fi
%Type = Article
\bibitem[{Kipf and Welling(2017)}]{kipf_semi-supervised_2017}
\bibinfo{author}{T.~N. Kipf}, \bibinfo{author}{M.~Welling},
\newblock \bibinfo{title}{Semi-{Supervised} {Classification} with {Graph}
  {Convolutional} {Networks}},
\newblock \bibinfo{journal}{ICLR}  (\bibinfo{year}{2017}).
%Type = Article
\bibitem[{Velickovic et~al.(2018)Velickovic, Cucurull, Casanova, Romero, Liò,
  and Bengio}]{velickovic_graph_2018}
\bibinfo{author}{P.~Velickovic}, \bibinfo{author}{G.~Cucurull},
  \bibinfo{author}{A.~Casanova}, \bibinfo{author}{A.~Romero},
  \bibinfo{author}{P.~Liò}, \bibinfo{author}{Y.~Bengio},
\newblock \bibinfo{title}{Graph {Attention} {Networks}},
\newblock \bibinfo{journal}{ICLR}  (\bibinfo{year}{2018}).
%Type = Inproceedings
\bibitem[{Abu-El-Haija et~al.(2019)Abu-El-Haija, Perozzi, Kapoor, Harutyunyan,
  Alipourfard, Lerman, Steeg, and Galstyan}]{abu-el-haija_mixhop_2019}
\bibinfo{author}{S.~Abu-El-Haija}, \bibinfo{author}{B.~Perozzi},
  \bibinfo{author}{A.~Kapoor}, \bibinfo{author}{H.~Harutyunyan},
  \bibinfo{author}{N.~Alipourfard}, \bibinfo{author}{K.~Lerman},
  \bibinfo{author}{G.~V. Steeg}, \bibinfo{author}{A.~Galstyan},
\newblock \bibinfo{title}{{MixHop}: {Higher}-{Order} {Graph} {Convolutional}
  {Architectures} via {Sparsified} {Neighborhood} {Mixing}},
\newblock in: \bibinfo{booktitle}{{ICML}}, \bibinfo{year}{2019}.
%Type = Article
\bibitem[{Chen et~al.(2020)Chen, Wei, Huang, Ding, and Li}]{chen_simple_2020}
\bibinfo{author}{M.~Chen}, \bibinfo{author}{Z.~Wei},
  \bibinfo{author}{Z.~Huang}, \bibinfo{author}{B.~Ding},
  \bibinfo{author}{Y.~Li},
\newblock \bibinfo{title}{Simple and {Deep} {Graph} {Convolutional}
  {Networks}},
\newblock \bibinfo{journal}{ICML}  (\bibinfo{year}{2020}).
%Type = Article
\bibitem[{Wang and Derr(2021)}]{wang_tree_2021}
\bibinfo{author}{Y.~Wang}, \bibinfo{author}{T.~Derr},
\newblock \bibinfo{title}{Tree {Decomposed} {Graph} {Neural} {Network}},
\newblock \bibinfo{journal}{CIKM}  (\bibinfo{year}{2021}).
%Type = Inproceedings
\bibitem[{Ying et~al.(2018)Ying, He, Chen, Eksombatchai, Hamilton, and
  Leskovec}]{ying_graph_2018}
\bibinfo{author}{R.~Ying}, \bibinfo{author}{R.~He}, \bibinfo{author}{K.~Chen},
  \bibinfo{author}{P.~Eksombatchai}, \bibinfo{author}{W.~L. Hamilton},
  \bibinfo{author}{J.~Leskovec},
\newblock \bibinfo{title}{Graph {Convolutional} {Neural} {Networks} for
  {Web}-{Scale} {Recommender} {Systems}},
\newblock in: \bibinfo{booktitle}{{KDD} '18}, \bibinfo{year}{2018}.
%Type = Article
\bibitem[{van~den Berg et~al.(2017)van~den Berg, Kipf, and
  Welling}]{van_den_berg_graph_2017}
\bibinfo{author}{R.~van~den Berg}, \bibinfo{author}{T.~Kipf},
  \bibinfo{author}{M.~Welling},
\newblock \bibinfo{title}{Graph {Convolutional} {Matrix} {Completion}},
\newblock \bibinfo{journal}{ArXiv} \bibinfo{volume}{abs/1706.02263}
  (\bibinfo{year}{2017}).
%Type = Article
\bibitem[{Chami et~al.(2019)Chami, Ying, Ré, and
  Leskovec}]{chami_hyperbolic_2019}
\bibinfo{author}{I.~Chami}, \bibinfo{author}{R.~Ying},
  \bibinfo{author}{C.~Ré}, \bibinfo{author}{J.~Leskovec},
\newblock \bibinfo{title}{Hyperbolic {Graph} {Convolutional} {Neural}
  {Networks}},
\newblock \bibinfo{journal}{NeurIPS}  (\bibinfo{year}{2019}).
%Type = Inproceedings
\bibitem[{Ying et~al.(2018)Ying, You, Morris, Ren, Hamilton, and
  Leskovec}]{ying_hierarchical_2018}
\bibinfo{author}{R.~Ying}, \bibinfo{author}{J.~You},
  \bibinfo{author}{C.~Morris}, \bibinfo{author}{X.~Ren}, \bibinfo{author}{W.~L.
  Hamilton}, \bibinfo{author}{J.~Leskovec},
\newblock \bibinfo{title}{Hierarchical {Graph} {Representation} {Learning} with
  {Differentiable} {Pooling}},
\newblock {NIPS}'18, \bibinfo{year}{2018}, pp. \bibinfo{pages}{4805--4815}.
%Type = Inproceedings
\bibitem[{Zhang et~al.(2018)Zhang, Cui, Neumann, and
  Chen}]{zhang_end--end_2018}
\bibinfo{author}{M.~Zhang}, \bibinfo{author}{Z.~Cui},
  \bibinfo{author}{M.~Neumann}, \bibinfo{author}{Y.~Chen},
\newblock \bibinfo{title}{An {End}-to-{End} {Deep} {Learning} {Architecture}
  for {Graph} {Classification}},
\newblock in: \bibinfo{booktitle}{{AAAI}}, \bibinfo{year}{2018}.
%Type = Inproceedings
\bibitem[{Gilmer et~al.(2017)Gilmer, Schoenholz, Riley, Vinyals, and
  Dahl}]{gilmer_neural_2017}
\bibinfo{author}{J.~Gilmer}, \bibinfo{author}{S.~S. Schoenholz},
  \bibinfo{author}{P.~F. Riley}, \bibinfo{author}{O.~Vinyals},
  \bibinfo{author}{G.~E. Dahl},
\newblock \bibinfo{title}{Neural {Message} {Passing} for {Quantum}
  {Chemistry}},
\newblock in: \bibinfo{booktitle}{Proceedings of the 34th {International}
  {Conference} on {Machine} {Learning} - {Volume} 70}, {ICML}'17,
  \bibinfo{year}{2017}.
%Type = Article
\bibitem[{Madhawa et~al.(2019)Madhawa, Ishiguro, Nakago, and
  Abe}]{madhawa_graphnvp_2019}
\bibinfo{author}{K.~Madhawa}, \bibinfo{author}{K.~Ishiguro},
  \bibinfo{author}{K.~Nakago}, \bibinfo{author}{M.~Abe},
\newblock \bibinfo{title}{{GraphNVP}: {An} {Invertible} {Flow} {Model} for
  {Generating} {Molecular} {Graphs}},
\newblock \bibinfo{journal}{arXiv:1905.11600}  (\bibinfo{year}{2019}).
%Type = Article
\bibitem[{Maurya et~al.(2021)Maurya, Liu, and Murata}]{maurya_graph_2021}
\bibinfo{author}{S.~K. Maurya}, \bibinfo{author}{X.~Liu},
  \bibinfo{author}{T.~Murata},
\newblock \bibinfo{title}{Graph {Neural} {Networks} for {Fast} {Node} {Ranking}
  {Approximation}},
\newblock \bibinfo{journal}{ACM Transactions on Knowledge Discovery from Data}
  \bibinfo{volume}{15} (\bibinfo{year}{2021}) \bibinfo{pages}{78:1--78:32}.
%Type = Inproceedings
\bibitem[{Fan et~al.(2019)Fan, Zeng, Ding, Chen, Sun, and
  Liu}]{fan_learning_2019}
\bibinfo{author}{C.~Fan}, \bibinfo{author}{L.~Zeng}, \bibinfo{author}{Y.~Ding},
  \bibinfo{author}{M.~Chen}, \bibinfo{author}{Y.~Sun},
  \bibinfo{author}{Z.~Liu},
\newblock \bibinfo{title}{Learning to {Identify} {High} {Betweenness}
  {Centrality} {Nodes} from {Scratch}: {A} {Novel} {Graph} {Neural} {Network}
  {Approach}},
\newblock in: \bibinfo{booktitle}{Proceedings of the 28th {ACM} {International}
  {Conference} on {Information} and {Knowledge} {Management}}, {CIKM} '19,
  \bibinfo{publisher}{ACM}, \bibinfo{address}{New York, NY, USA},
  \bibinfo{year}{2019}, pp. \bibinfo{pages}{559--568}.
%Type = Inproceedings
\bibitem[{Marcheggiani and Titov(2017)}]{marcheggiani_encoding_2017}
\bibinfo{author}{D.~Marcheggiani}, \bibinfo{author}{I.~Titov},
\newblock \bibinfo{title}{Encoding {Sentences} with {Graph} {Convolutional}
  {Networks} for {Semantic} {Role} {Labeling}},
\newblock \bibinfo{publisher}{ACL}, \bibinfo{year}{2017}, pp.
  \bibinfo{pages}{1506--1515}.
%Type = Article
\bibitem[{Wu et~al.(2019)Wu, Pan, Chen, Long, Zhang, and
  Yu}]{wu_comprehensive_2019}
\bibinfo{author}{Z.~Wu}, \bibinfo{author}{S.~Pan}, \bibinfo{author}{F.~Chen},
  \bibinfo{author}{G.~Long}, \bibinfo{author}{C.~Zhang}, \bibinfo{author}{P.~S.
  Yu},
\newblock \bibinfo{title}{A {Comprehensive} {Survey} on {Graph} {Neural}
  {Networks}},
\newblock \bibinfo{journal}{arXiv:1901.00596 [cs, stat]}
  (\bibinfo{year}{2019}).
%Type = Incollection
\bibitem[{Hamilton et~al.(2017)Hamilton, Ying, and
  Leskovec}]{hamilton_inductive_2017}
\bibinfo{author}{W.~Hamilton}, \bibinfo{author}{Z.~Ying},
  \bibinfo{author}{J.~Leskovec},
\newblock \bibinfo{title}{Inductive {Representation} {Learning} on {Large}
  {Graphs}},
\newblock in: \bibinfo{booktitle}{{NIPS}}, \bibinfo{year}{2017}, pp.
  \bibinfo{pages}{1024--1034}.
%Type = Inproceedings
\bibitem[{Chen et~al.(2018)Chen, Ma, and Xiao}]{chen_fastgcn_2018}
\bibinfo{author}{J.~Chen}, \bibinfo{author}{T.~Ma}, \bibinfo{author}{C.~Xiao},
\newblock \bibinfo{title}{{FastGCN}: {Fast} {Learning} with {Graph}
  {Convolutional} {Networks} via {Importance} {Sampling}},
\newblock in: \bibinfo{booktitle}{{ICLR}}, \bibinfo{year}{2018}.
%Type = Misc
\bibitem[{Klicpera et~al.(2018)Klicpera, Bojchevski, and
  Günnemann}]{klicpera_predict_2018}
\bibinfo{author}{J.~Klicpera}, \bibinfo{author}{A.~Bojchevski},
  \bibinfo{author}{S.~Günnemann}, \bibinfo{title}{Predict then {Propagate}:
  {Combining} neural networks with personalized pagerank for classification on
  graphs}, \bibinfo{year}{2018}.
%Type = Inproceedings
\bibitem[{Jin et~al.(2021)Jin, Derr, Wang, Ma, Liu, and Tang}]{jin_node_2021}
\bibinfo{author}{W.~Jin}, \bibinfo{author}{T.~Derr}, \bibinfo{author}{Y.~Wang},
  \bibinfo{author}{Y.~Ma}, \bibinfo{author}{Z.~Liu}, \bibinfo{author}{J.~Tang},
\newblock \bibinfo{title}{Node {Similarity} {Preserving} {Graph}
  {Convolutional} {Networks}},
\newblock {WSDM} '21, \bibinfo{publisher}{Association for Computing Machinery},
  \bibinfo{year}{2021}, pp. \bibinfo{pages}{148--156}.
%Type = Inproceedings
\bibitem[{Wu et~al.(2019)Wu, Souza, Zhang, Fifty, Yu, and
  Weinberger}]{wu_simplifying_2019}
\bibinfo{author}{F.~Wu}, \bibinfo{author}{A.~H. Souza},
  \bibinfo{author}{T.~Zhang}, \bibinfo{author}{C.~Fifty},
  \bibinfo{author}{T.~Yu}, \bibinfo{author}{K.~Q. Weinberger},
\newblock \bibinfo{title}{Simplifying {Graph} {Convolutional} {Networks}},
\newblock in: \bibinfo{booktitle}{{ICML}}, \bibinfo{year}{2019}.
%Type = Inproceedings
\bibitem[{Rong et~al.(2020)Rong, Huang, Xu, and Huang}]{rong_dropedge_2020}
\bibinfo{author}{Y.~Rong}, \bibinfo{author}{W.~Huang}, \bibinfo{author}{T.~Xu},
  \bibinfo{author}{J.~Huang},
\newblock \bibinfo{title}{{DropEdge}: {Towards} {Deep} {Graph} {Convolutional}
  {Networks} on {Node} {Classification}},
\newblock in: \bibinfo{booktitle}{{ICLR}}, \bibinfo{year}{2020}.
%Type = Article
\bibitem[{Zhu et~al.(2020)Zhu, Yan, Zhao, Heimann, Akoglu, and
  Koutra}]{zhu_beyond_2020}
\bibinfo{author}{J.~Zhu}, \bibinfo{author}{Y.~Yan}, \bibinfo{author}{L.~Zhao},
  \bibinfo{author}{M.~Heimann}, \bibinfo{author}{L.~Akoglu},
  \bibinfo{author}{D.~Koutra},
\newblock \bibinfo{title}{Beyond {Homophily} in {Graph} {Neural} {Networks}:
  {Current} {Limitations} and {Effective} {Designs}},
\newblock \bibinfo{journal}{NeurIPS} \bibinfo{volume}{33}
  (\bibinfo{year}{2020}).
%Type = Inproceedings
\bibitem[{Zhu et~al.(2021)Zhu, Rossi, Rao, Mai, Lipka, Ahmed, and
  Koutra}]{zhu_graph_2021}
\bibinfo{author}{J.~Zhu}, \bibinfo{author}{R.~A. Rossi}, \bibinfo{author}{A.~B.
  Rao}, \bibinfo{author}{T.~Mai}, \bibinfo{author}{N.~Lipka},
  \bibinfo{author}{N.~Ahmed}, \bibinfo{author}{D.~Koutra},
\newblock \bibinfo{title}{Graph {Neural} {Networks} with {Heterophily}},
\newblock in: \bibinfo{booktitle}{{AAAI}}, \bibinfo{year}{2021}.
%Type = Inproceedings
\bibitem[{Bo et~al.(2021)Bo, Wang, Shi, and Shen}]{bo_beyond_2021}
\bibinfo{author}{D.~Bo}, \bibinfo{author}{X.~Wang}, \bibinfo{author}{C.~Shi},
  \bibinfo{author}{H.~Shen},
\newblock \bibinfo{title}{Beyond {Low}-frequency {Information} in {Graph}
  {Convolutional} {Networks}},
\newblock in: \bibinfo{booktitle}{{AAAI}}, \bibinfo{year}{2021}.
%Type = Article
\bibitem[{Maurya et~al.(2021)Maurya, Liu, and Murata}]{maurya_improving_2021}
\bibinfo{author}{S.~K. Maurya}, \bibinfo{author}{X.~Liu},
  \bibinfo{author}{T.~Murata},
\newblock \bibinfo{title}{Improving {Graph} {Neural} {Networks} with {Simple}
  {Architecture} {Design}},
\newblock \bibinfo{journal}{arXiv:2105.07634}  (\bibinfo{year}{2021}).
%Type = Article
\bibitem[{Frasca et~al.(2020)Frasca, Rossi, Eynard, Chamberlain, Bronstein, and
  Monti}]{frasca_sign_2020}
\bibinfo{author}{F.~Frasca}, \bibinfo{author}{E.~Rossi},
  \bibinfo{author}{D.~Eynard}, \bibinfo{author}{B.~Chamberlain},
  \bibinfo{author}{M.~Bronstein}, \bibinfo{author}{F.~Monti},
\newblock \bibinfo{title}{{SIGN}: {Scalable} {Inception} {Graph} {Neural}
  {Networks}},
\newblock \bibinfo{journal}{arXiv:2004.11198 [cs, stat]}
  (\bibinfo{year}{2020}).
%Type = Article
\bibitem[{Tang et~al.(2014)Tang, Alelyani, and Liu}]{tang_feature_2014}
\bibinfo{author}{J.~Tang}, \bibinfo{author}{S.~Alelyani},
  \bibinfo{author}{H.~Liu},
\newblock \bibinfo{title}{Feature selection for classification: {A} review},
\newblock \bibinfo{journal}{Data Classification: Algorithms and Applications}
  (\bibinfo{year}{2014}) \bibinfo{pages}{37--64}.
%Type = Article
\bibitem[{Li et~al.(2017)Li, Cheng, Wang, Morstatter, Trevino, Tang, and
  Liu}]{li_feature_2017}
\bibinfo{author}{J.~Li}, \bibinfo{author}{K.~Cheng}, \bibinfo{author}{S.~Wang},
  \bibinfo{author}{F.~Morstatter}, \bibinfo{author}{R.~P. Trevino},
  \bibinfo{author}{J.~Tang}, \bibinfo{author}{H.~Liu},
\newblock \bibinfo{title}{Feature {Selection}: {A} {Data} {Perspective}},
\newblock \bibinfo{journal}{ACM Computing Surveys} \bibinfo{volume}{50}
  (\bibinfo{year}{2017}) \bibinfo{pages}{94:1--94:45}.
%Type = Article
\bibitem[{Chandrashekar and Sahin(2014)}]{chandrashekar_survey_2014}
\bibinfo{author}{G.~Chandrashekar}, \bibinfo{author}{F.~Sahin},
\newblock \bibinfo{title}{A survey on feature selection methods},
\newblock \bibinfo{journal}{Computers \& Electrical Engineering}
  \bibinfo{volume}{40} (\bibinfo{year}{2014}) \bibinfo{pages}{16--28}.
%Type = Article
\bibitem[{Chien et~al.(2021)Chien, Peng, Li, and
  Milenkovic}]{chien_adaptive_2021}
\bibinfo{author}{E.~Chien}, \bibinfo{author}{J.~Peng}, \bibinfo{author}{P.~Li},
  \bibinfo{author}{O.~Milenkovic},
\newblock \bibinfo{title}{Adaptive {Universal} {Generalized} {PageRank} {Graph}
  {Neural} {Network}},
\newblock \bibinfo{journal}{ICLR}  (\bibinfo{year}{2021}).
%Type = Article
\bibitem[{Pei et~al.(2020)Pei, Wei, Chang, Lei, and Yang}]{pei_geom-gcn_2020}
\bibinfo{author}{H.~Pei}, \bibinfo{author}{B.~Wei}, \bibinfo{author}{K.~Chang},
  \bibinfo{author}{Y.~Lei}, \bibinfo{author}{B.~Yang},
\newblock \bibinfo{title}{Geom-{GCN}: {Geometric} {Graph} {Convolutional}
  {Networks}},
\newblock \bibinfo{journal}{ICLR}  (\bibinfo{year}{2020}).
%Type = Article
\bibitem[{Suresh et~al.(2021)Suresh, Budde, Neville, Li, and
  Ma}]{suresh_breaking_2021}
\bibinfo{author}{S.~Suresh}, \bibinfo{author}{V.~Budde},
  \bibinfo{author}{J.~Neville}, \bibinfo{author}{P.~Li},
  \bibinfo{author}{J.~Ma},
\newblock \bibinfo{title}{Breaking the {Limit} of {Graph} {Neural} {Networks}
  by {Improving} the {Assortativity} of {Graphs} with {Local} {Mixing}
  {Patterns}},
\newblock \bibinfo{journal}{arXiv:2106.06586 [cs]}  (\bibinfo{year}{2021}).
%Type = Article
\bibitem[{NT et~al.(2020)NT, Maehara, and Murata}]{nt_stacked_2020}
\bibinfo{author}{H.~NT}, \bibinfo{author}{T.~Maehara},
  \bibinfo{author}{T.~Murata},
\newblock \bibinfo{title}{Stacked {Graph} {Filter}},
\newblock \bibinfo{journal}{arXiv:2011.10988 [cs]}  (\bibinfo{year}{2020}).
%Type = Article
\bibitem[{Berberidis et~al.(2019)Berberidis, Nikolakopoulos, and
  Giannakis}]{berberidis_adaptive_2019}
\bibinfo{author}{D.~Berberidis}, \bibinfo{author}{A.~N. Nikolakopoulos},
  \bibinfo{author}{G.~B. Giannakis},
\newblock \bibinfo{title}{Adaptive {Diffusions} for {Scalable} {Learning}
  {Over} {Graphs}},
\newblock \bibinfo{journal}{IEEE Transactions on Signal Processing}
  \bibinfo{volume}{67} (\bibinfo{year}{2019}) \bibinfo{pages}{1307--1321}.
%Type = Article
\bibitem[{Defferrard et~al.(2016)Defferrard, Bresson, and
  Vandergheynst}]{defferrard_convolutional_2016}
\bibinfo{author}{M.~Defferrard}, \bibinfo{author}{X.~Bresson},
  \bibinfo{author}{P.~Vandergheynst},
\newblock \bibinfo{title}{Convolutional {Neural} {Networks} on {Graphs} with
  {Fast} {Localized} {Spectral} {Filtering}},
\newblock \bibinfo{journal}{arXiv:1606.09375}  (\bibinfo{year}{2016}).
%Type = Inproceedings
\bibitem[{Xu et~al.(2018)Xu, Li, Tian, Sonobe, Kawarabayashi, and
  Jegelka}]{xu_representation_2018}
\bibinfo{author}{K.~Xu}, \bibinfo{author}{C.~Li}, \bibinfo{author}{Y.~Tian},
  \bibinfo{author}{T.~Sonobe}, \bibinfo{author}{K.-i. Kawarabayashi},
  \bibinfo{author}{S.~Jegelka},
\newblock \bibinfo{title}{Representation {Learning} on {Graphs} with {Jumping}
  {Knowledge} {Networks}},
\newblock \bibinfo{publisher}{PMLR}, \bibinfo{year}{2018}.
%Type = Inproceedings
\bibitem[{Li et~al.(2021)Li, Müller, Ghanem, and Koltun}]{li_training_2021}
\bibinfo{author}{G.~Li}, \bibinfo{author}{M.~Müller},
  \bibinfo{author}{B.~Ghanem}, \bibinfo{author}{V.~Koltun},
\newblock \bibinfo{title}{Training {Graph} {Neural} {Networks} with 1000
  {Layers}},
\newblock in: \bibinfo{booktitle}{{ICML}}, \bibinfo{year}{2021}.
%Type = Article
\bibitem[{Godwin et~al.(2021)Godwin, Schaarschmidt, Gaunt, Sanchez-Gonzalez,
  Rubanova, Veličković, Kirkpatrick, and Battaglia}]{godwin_very_2021}
\bibinfo{author}{J.~Godwin}, \bibinfo{author}{M.~Schaarschmidt},
  \bibinfo{author}{A.~Gaunt}, \bibinfo{author}{A.~Sanchez-Gonzalez},
  \bibinfo{author}{Y.~Rubanova}, \bibinfo{author}{P.~Veličković},
  \bibinfo{author}{J.~Kirkpatrick}, \bibinfo{author}{P.~Battaglia},
\newblock \bibinfo{title}{Very {Deep} {Graph} {Neural} {Networks} {Via} {Noise}
  {Regularisation}},
\newblock \bibinfo{journal}{arXiv:2106.07971 [cs]}  (\bibinfo{year}{2021}).
%Type = Article
\bibitem[{Ma et~al.(2021)Ma, Liu, Shah, and Tang}]{ma_is_2021}
\bibinfo{author}{Y.~Ma}, \bibinfo{author}{X.~Liu}, \bibinfo{author}{N.~Shah},
  \bibinfo{author}{J.~Tang},
\newblock \bibinfo{title}{Is {Homophily} a {Necessity} for {Graph} {Neural}
  {Networks}?},
\newblock \bibinfo{journal}{arXiv:2106.06134 [cs, stat]}
  (\bibinfo{year}{2021}).
%Type = Article
\bibitem[{Sen et~al.(2008)Sen, Namata, Bilgic, Getoor, Gallagher, and
  Eliassi-Rad}]{sen_collective_2008}
\bibinfo{author}{P.~Sen}, \bibinfo{author}{G.~Namata},
  \bibinfo{author}{M.~Bilgic}, \bibinfo{author}{L.~Getoor},
  \bibinfo{author}{B.~Gallagher}, \bibinfo{author}{T.~Eliassi-Rad},
\newblock \bibinfo{title}{Collective {Classification} in {Network} {Data}},
\newblock \bibinfo{journal}{AI Mag.}  (\bibinfo{year}{2008}).
%Type = Inproceedings
\bibitem[{Tang et~al.(2009)Tang, Sun, Wang, and Yang}]{tang_social_2009}
\bibinfo{author}{J.~Tang}, \bibinfo{author}{J.~Sun}, \bibinfo{author}{C.~Wang},
  \bibinfo{author}{Z.~Yang},
\newblock \bibinfo{title}{Social influence analysis in large-scale networks},
\newblock {KDD} '09, \bibinfo{publisher}{Association for Computing Machinery},
  \bibinfo{year}{2009}, pp. \bibinfo{pages}{807--816}.
%Type = Article
\bibitem[{Rozemberczki et~al.(2020)Rozemberczki, Allen, and
  Sarkar}]{rozemberczki_multi-scale_2020}
\bibinfo{author}{B.~Rozemberczki}, \bibinfo{author}{C.~Allen},
  \bibinfo{author}{R.~Sarkar},
\newblock \bibinfo{title}{Multi-scale {Attributed} {Node} {Embedding}},
\newblock \bibinfo{journal}{arXiv:1909.13021 [cs, stat]}
  (\bibinfo{year}{2020}).

\end{thebibliography}


% Author biography without author photo.
% Author biography. Author biography. Author biography.
% Author biography. Author biography. Author biography.
% Author biography. Author biography. Author biography.
% Author biography. Author biography. Author biography.
% Author biography. Author biography. Author biography.
% Author biography. Author biography. Author biography.
% Author biography. Author biography. Author biography.
% Author biography. Author biography. Author biography.
% Author biography. Author biography. Author biography.

% \bio{figs/pic1}
% Author biography with author photo.
% Author biography. Author biography. Author biography.
% Author biography. Author biography. Author biography.
% Author biography. Author biography. Author biography.
% Author biography. Author biography. Author biography.
% Author biography. Author biography. Author biography.
% Author biography. Author biography. Author biography.
% Author biography. Author biography. Author biography.
% Author biography. Author biography. Author biography.
% Author biography. Author biography. Author biography.

% \bio{figs/pic1}
% Author biography with author photo.
% Author biography. Author biography. Author biography.
% Author biography. Author biography. Author biography.
% Author biography. Author biography. Author biography.
% Author biography. Author biography. Author biography.

\end{document}